\documentclass[journal]{IEEEtran}

\IEEEoverridecommandlockouts     

\usepackage{graphicx}
\usepackage{caption}
\usepackage{subfig}
\usepackage{amsmath}

\usepackage{multirow}

\title{\LARGE \bf
An Automated Analysis Framework \\ for Trajectory Datasets
}

\author{Christoph Glasmacher$^{1}$, Robert Krajewski$^{2}$ and Lutz Eckstein$^{1}$

\thanks{$^{1}$ The authors are with the Institute for Automotive Engineering, RWTH Aachen University (Aachen, Germany). E-mail:\{christoph.glasmacher, lutz.eckstein\}@ika.rwth-aachen.de}
\thanks{$^{2}$Robert Krajewski is with the RWTH Aachen University. E-mail: robert.krajewski@rwth-aachen.de}%
}

\begin{document}

\let\oldtwocolumn\twocolumn
\renewcommand\twocolumn[1][]{%
	\oldtwocolumn[{#1}{
		\begin{center}
    		%\vspace*{-1cm}
			\includegraphics[width=0.8\textwidth]{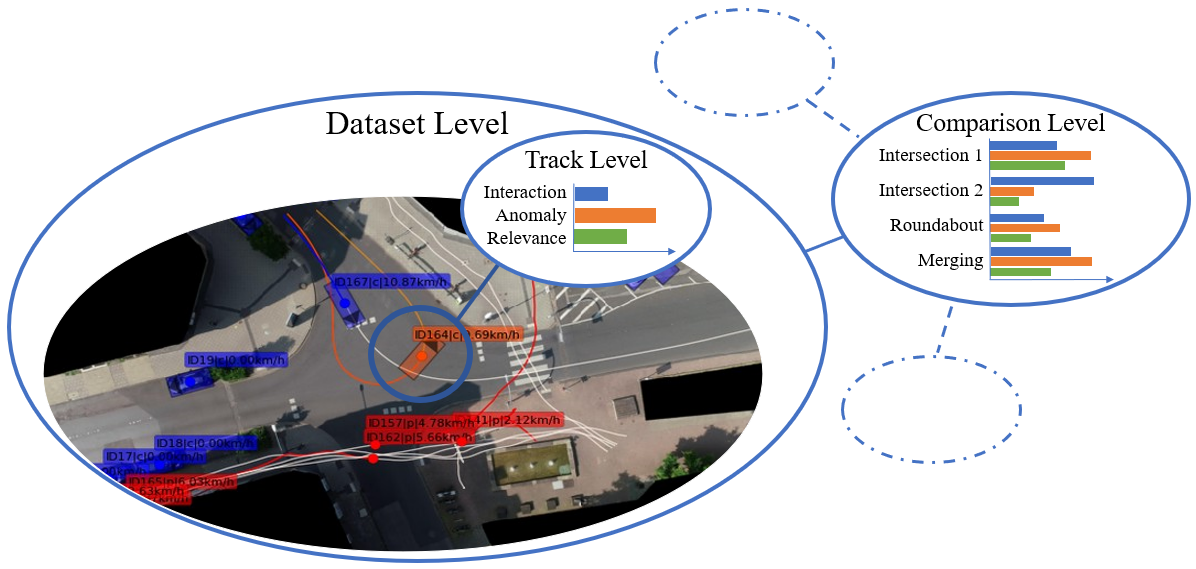}
			\captionof{figure}{Schematic description of the proposed analysis framework. The behavior of each vehicle is analyzed in each timestep. According to its characteristics, interaction, anomaly and relevance are calculated. These values are accumulated on a track and dataset level for comparison between different datasets.}
			\label{fig:headliner}
		\end{center}
	}]
}

\maketitle
\thispagestyle{empty}
\pagestyle{empty}

%%%%%%%%%%%%%%%%%%%%%%%%%%%%%%%%%%%%%%%%%%%%%%%%%%%%%%%%%%%%%%%%%%%%%%%%%%%%%%%%
\begin{abstract}

Trajectory datasets of road users have become more important in the last years for safety validation of highly automated vehicles. Several naturalistic trajectory datasets with each more than 10~000 tracks were released and others will follow. Considering this amount of data, it is necessary to be able to compare these datasets in-depth with ease to get an overview. By now, the datasets' own provided information is mainly limited to meta-data and qualitative descriptions which are mostly not consistent with other datasets. This is insufficient for users to differentiate the emerging datasets for application-specific selection. Therefore, an automated analysis framework is proposed in this work. Starting with analyzing individual tracks, fourteen elementary characteristics, so-called detection types, are derived and used as the base of this framework. To describe each traffic scenario precisely, the detections are subdivided into common metrics, clustering methods and anomaly detection. Those are combined using a modular approach. The detections are composed into new scores to describe three defined attributes of each track data quantitatively: interaction, anomaly and relevance. These three scores are calculated hierarchically for different abstract layers to provide an overview not just between datasets but also for tracks, spatial regions and individual situations. So, an objective comparison between datasets can be realized. Furthermore, it can help to get a deeper understanding of the recorded infrastructure and its effect on road user behavior. To test the validity of the framework, a study is conducted to compare the scores with human perception. Additionally, several datasets are compared.

\end{abstract}

%%%%%%%%%%%%%%%%%%%%%%%%%%%%%%%%%%%%%%%%%%%%%%%%%%%%%%%%%%%%%%%%%%%%%%%%%%%%%%%%
\section{INTRODUCTION}

The analysis of traffic flow and road user behavior become increasingly important for both, infrastructure design and driving applications. Especially, to ensure a high standard of traffic safety with increasing automation, verification and validation have taken a central role in the development of automated driving functions.
Since validation by real tests exclusively is neither economically viable nor feasible in a reasonable time, software-based tests play a decisive role~\cite{importance_of_software_tests}. The data required for these tests cannot be generated completely synthetically. For a broad coverage of scenarios, the use of real recorded data is essential, so that a system can be confronted with diverse situations like highly interactive or unusual road user behavior~\cite{highD}. 
For this purpose, multiple trajectory datasets have been published in the last years~\cite{Dataset_revisit_Li}. Thereby, naturalistic trajectories of road users recorded from a bird's-eye view form an important class of datasets. 
With this kind of dataset, the interplay of road users at individual intersections or other locally bounded infrastructures can be detected and analyzed within relatively fixed boundary conditions~\cite{pNEUMA}. 
The increasing amount of recorded data allows the user to improve testing and verification of systems and applications. But to use this data, it is crucial that the enormous amount can be overlooked.
For this purpose, datasets are typically accompanied by a separate publication. These mostly focus on the methodology used for the trajectory extraction and not on their quality~\cite{inD}. Usually, little more about the given trajectories is provided than rudimentary information. For example, the \textit{highD} paper \cite{highD} solely makes statements about the behavior of cut-in scenarios and velocity distribution. In addition, the respective dataset is usually described qualitatively, but characteristics of trajectories like their velocities, paths or interactions are rarely considered empirically~\cite{pNEUMA}.
There are hardly any quantitative statements about the interaction and none about the anomaly and relevance of tracks or scenarios. Also, not every dataset is well suited for every application as very few interactions take place or nearly all road users behave according to traffic rules. Therefore, experts typically have to review datasets manually and score them subjectively due to their interesting situations like critical, complex or unusual ones. Especially those real road user constellations that might not be considered in simulations, have to be used to validate systems. In order to improve the feasibility and efficiency of this process, automatically derived scores are needed that match the perception of the experts~\cite{driving_factors}.

For that purpose, as the main contribution, an automated analysis framework for trajectory datasets is proposed in this paper. 
We define an interaction score, an anomaly score as behavior that highly deviates from the typical behavior in the recording and a relevance score as a combination of interaction and anomaly.
The hierarchical framework evaluates the three scores on four different levels of abstraction of the dataset (\ref{sec:methodology}).
Accumulating these scores over the abstract levels enables the user not only to evaluate individual tracks but also compare complete datasets (see Fig.~\ref{fig:headliner}).

After the presentation of the analysis framework, we compare it with experts and other algorithms for validation. Since there is no such absolute definition of interaction, a survey is set up to test the correlation between score and perception of experts.
Additionally, the framework is compared to the scheme of the \textit{INTERACTION} dataset~\cite{INTERACTION_dataset} that tries to measure interaction using different metrics.
Subsequently, further functions are presented and application fields are shown.

\section{RELATED WORK}

As a basis for the analysis framework developed within this paper, we provide an overview of previous approaches for evaluating traffic scenarios, as well as an overview of their specific use for recent trajectory datasets.

\subsection{Assessment Systems for Traffic Scenarios}
\label{sec:relatedwork_assesment}

Research on traffic scenario analysis began long before the publication of large trajectory datasets. Already in the 1970s, metrics for the description of conflicts between vehicles were presented for accident research:
\begin{itemize}
    \item \textbf{Time to collision} (TTC): The metric describes the time to a potential crash of two road users assuming an unchanged driving state~\cite{TTC_Hayward}.
    \item \textbf{Time headway} (THW): It measures the time it takes for a following vehicle to reach the current position of the followed vehicle, assuming constant speed~\cite{THW_Vogel}.
    \item \textbf{Deceleration rate to avoid a crash} (DRAC): The metric (\ref{eq:DRAC}) describes the minimum average delay ($a_{dec}$) of a road user to avoid an accident at given velocities (v) and distance between the vehicles ($\Delta x$)~\cite{DRAC_explanation}.
\end{itemize}
\begin{align}
	DRAC = \frac{(v_2 - v_1)^2}{2 \Delta x}
	\label{eq:DRAC}
\end{align}
Despite their development for accident research, they are still used today with regard to automated driving~\cite{usage_of_metrics_automated_driving, risk_measure_ttc}. At the same time, new metrics have been developed that focus on the application for automated vehicles.~\cite{INTERACTION_scheme}:
\begin{itemize}
    \item \textbf{Difference of time to conflict point} ($\Delta$TTCP): The $\Delta$TTCP describes the time difference in which two road users pass the same point (a point of conflict).
    \item \textbf{Waiting period} (WP): The metric describes the time a road user stands still with the intention to move on.
\end{itemize}
All of these metrics have in common that they are defined for linear scenarios or, in the case of the $\Delta$TTCP, require predefined \textit{Merging Points}~\cite{INTERACTION_scheme}, which makes the availability of map data necessary. 
Nevertheless, some metrics are frequently used due to their simplicity and have been validated in experiments~\cite{TTC_THW_experiment}. In the case of the THW, it is currently used in road traffic regulations~\cite{THW_Vogel}. However, most of these metrics can only be applied correctly to special scenarios like following scenarios e.g. on highways. 
Due to this kind of limitations, adaptations are needed for using and combining metrics in more complex scenarios.

In addition to metrics, additional approaches were pursued to describe characteristics: for example in the area of behavioral description~\cite{WAN20, Stanford_Drone_Dataset} and trajectory prediction~\cite{trajectory_prediction}. The focus is usually set on the search of patterns and the deviation from the rule in specific infrastructures like e.g. highways. To find these patterns, density-based approaches are used mainly for anomaly detection of trajectories \cite{framework_anomaly_detection, anomaly_detection_score}. Therefore, clustering methods~\cite{trajectory_clustering_CHO} or neuronal networks~\cite{trajectory_neural_network} are set up to group attributes and find unusual data. 
Besides these density-based methods, especially trajectory prediction methods combine various influences to determine the best possible trajectory based on complex cost functions including e.g. safety, visibility and comfort. For example~\cite{WAN15} uses an approach based on road conditions, driver behavior and vehicle state. It sums these up by generating a potential field through which the optimal route can be found. This approach allows the combination of different input parameters \cite{vehicle_limit_BEN15, NACHIKET18}. Based on these methods, several approaches were developed in the last few years, but almost none of them is used for actual dataset evaluation (\ref{sec:relatedwork_datasets}).

\subsection{Trajectory Datasets}
\label{sec:relatedwork_datasets}

Trajectory datasets have become increasingly relevant in recent years. This has resulted in a large number of public datasets~\cite{Dataset_revisit_Li}. In general, a distinction can be made between the utilization of offboard and onboard sensors in the dataset creation process. Onboard sensors can only perceive immediate surroundings around the vehicle and are limited by occlusions, so that they perceive traffic situations incompletely. However, because they are fixed to moving vehicles, datasets include a larger variety of infrastructures~\cite{range_onboard_sensors}. In contrast, offboard sensors are typically positioned stationary and thus record the behavioral variety at small sets of selected recording sites. Since they are not tied to vehicles, they typically have a better perspective and can record traffic situations over a long period~\cite{inD}. For example, in the \textit{inD} dataset~\cite{inD} intersections are observed and road users within are tracked. Thus, complex events including several traffic participants over time at fixed locations (e.g. intersections) can be recorded and analyzed in a next step.

With the increase in the number of recorded datasets, the number of relevant trajectory datasets recorded by offboard sensors is growing. In order to give the user an overview, information about the recorded road users is typically provided in a publication. However, these analyses are usually limited to basic quantitative parameters such as metrics or the number and type of traffic participants that are only analyzed independently~\cite{pNEUMA, inD, rounD}. In addition, a series of qualitative descriptions is usually carried out. Beyond that, only a few quantitative descriptions are provided. An example is the INTERACTION dataset~\cite{INTERACTION_dataset}, for which the interaction is analyzed using two predefined metrics~\cite{INTERACTION_scheme}. The authors compare similar datasets with those metrics. However, the employed scheme does not provide detailed information about individual scenarios, because metrics with similar values are just grouped. Those groups are only analyzed on the dataset level. This allows a rough comparison of the general interaction. However, it does not provide a detailed overview, nor a comprehensive interaction analysis of individual road users because only bilateral relations are analyzed but no more complex vehicle constellations.

\section{Methodology} %Guiding principle
\label{sec:methodology}

The framework proposed in this paper follows two principles: comprehensiveness and clarity. In order to achieve comprehensiveness, no specific use case is targeted to design the framework as unbiased as possible. So, it is universally applicable to find any kind of interesting situation. For this purpose, three characteristics are defined as essential scores of the framework: 
\begin{itemize}
    \item \textbf{Interaction}: Interaction describes the interplay of the road users among other traffic participants.
    \item \textbf{Anomaly}: Anomaly is driven by rarity~\cite{definition_anomaly}. The anomaly score highlights unusual phenomena or behavior.
    \item \textbf{Relevance}: Relevance is defined as the combination of interaction and anomaly.
\end{itemize}
While interaction and anomaly are generally known concepts, relevance is newly defined. It combines both points. Depending on the dataset usage, interactive scenarios may be less relevant if vehicles drive with constant velocity on a highway with other road users. A turning vehicle on an intersection might be unusual but it might not relevant until it interacts with another road user. Therefore, we define the product of both, interaction and anomaly, as relevance.
To ensure that the clarity is not lost despite the amount of information generated by applying these scores for a comprehensive analysis to every point in time, a hierarchical four-level approach is used.
In a first step, a modular set of rules is applied and characteristics (\textit{detections}) are extracted from the track data (chap.~\ref{sec:observed_scenarios}). These are evaluated and weighted to combine them to the three characteristic scores (chap.~\ref{chap:combining_scores}) on four layers (chap.~\ref{sec:hierachical_kombination}):
\begin{itemize}
    \item \textbf{Temporal punctual}: On the most basic level, the scores are defined for every road user in each timestep individually.
    \item \textbf{Track}: The temporal, punctual scores are merged to accumulate the scores for whole trajectories of road users.
    \item \textbf{Spatial region}: Besides the entity-focus, the scores can be derived for spatial regions of the infrastructure. Therefore, the behavior of road users within the region is considered.
    \item \textbf{Overall}: On the most abstract level, the scores of all detected road users are summed up to get a comprehensive picture of the whole dataset.
\end{itemize}
Considering the three types of scores on four abstract layers the user is given both: a high degree of clarity and the possibility of variable use.

\section{Traffic Situation Analysis}
\label{sec:observed_scenarios}

As the basis of the hierarchical framework, all road users including pedestrians and bicycles are observed individually at every timestep and are broken down to relevant attributes of its trajectory. Those attributes are called \textit{detections}. They form the foundation of the modular framework. These detections are derived by applying a set of criteria (\textit{detection types}). Fourteen different detection types in three thematic blocks are presented: vehicle relation indicators, individual vehicle state and context-related behavior. These are checked independently for each timestep and detections are assigned to both, the involved traffic participants and the associated infrastructure region. In order to provide a comprehensive picture, a minimum of dataset information needed for analysis is assumed. These requirements are kept general, so that neither country-specific restrictions nor a limitation to a certain type of scenario are considered.

\subsection{Dataset Requirements}
\label{sec:data_requirements}

According to chapter~\ref{sec:relatedwork_datasets}, the type of trajectory datasets can vary in their recording method. However, in order to generate a comprehensive analysis, the following restrictions are made with regard to the recording approach as well as the provided data to ensure that the datasets are comparable. Especially the recording approach offers the advantage of a temporally extended view of selected recording sites. So, complex traffic scenarios can be recorded within a limited spatial region. Those scenarios can go beyond the duration of stay of individual road users. This fact allows linking the infrastructure to road user behaviors.
Furthermore, for the processing by the proposed framework, datasets have to include three key elements:
	
\begin{itemize}
	\item \textbf{Trajectory data}: Information about velocity, acceleration, position and orientation of every road user for each frame
	\item \textbf{Meta data}: Information about road user dimensions and type to distinguish vulnerable road users from cars, as well as regulatory permission of region usage
	\item \textbf{Semantic map}: Information about the infrastructure including position and definition of characteristics such as lanes, possible driving paths, speed limits and walkways in a reasonable granularity (see OpenStreetMap~\cite{OpenStreetMap} or Lanelets~\cite{Lanelets})
\end{itemize}

The restrictions made are fulfilled by datasets like \textit{inD}~\cite{inD}, \textit{rounD}~\cite{rounD} or \textit{INTERACTION}~\cite{INTERACTION_dataset}). Thus, the request enables a detailed assessment so that datasets matching the requirements can be comprehensively mapped.

\subsection{Vehicle Relation Indicators}
\label{sec:detections_interaction}

An essential point in the evaluation of driving scenarios is the interaction between road users. These can sometimes become quite complex and can be both, temporally and spatially extended. Within our framework, complex scenarios are broken down into a combination of bilateral vehicle interactions to effectively take an unlimited number of traffic participants into account. To comprehensively map the individual interactions, several relation indicators (see Chap.~\ref{sec:relatedwork_assesment}) are used.
However, in order to be able to calculate these metrics automatically within more complex scenarios, the metrics have to be partially adapted because they are mostly designed for straight-line scenarios. Thus, the bilateral metrics are based on at least one extrapolation of a path with constant speed. Usually, the current speed of the object at the time of calculation is kept constant and the time gap is determined by the Euclidean distance to a conflict point with its interaction partner~\cite{calculation_metrics}. Due to the sometimes complex trajectories at intersections, this approach is revised: Instead of using only the Euclidean distance the actual driven path is considered in addition. 

\begin{figure}[thpb]
	\centering
	\includegraphics[width=\linewidth]{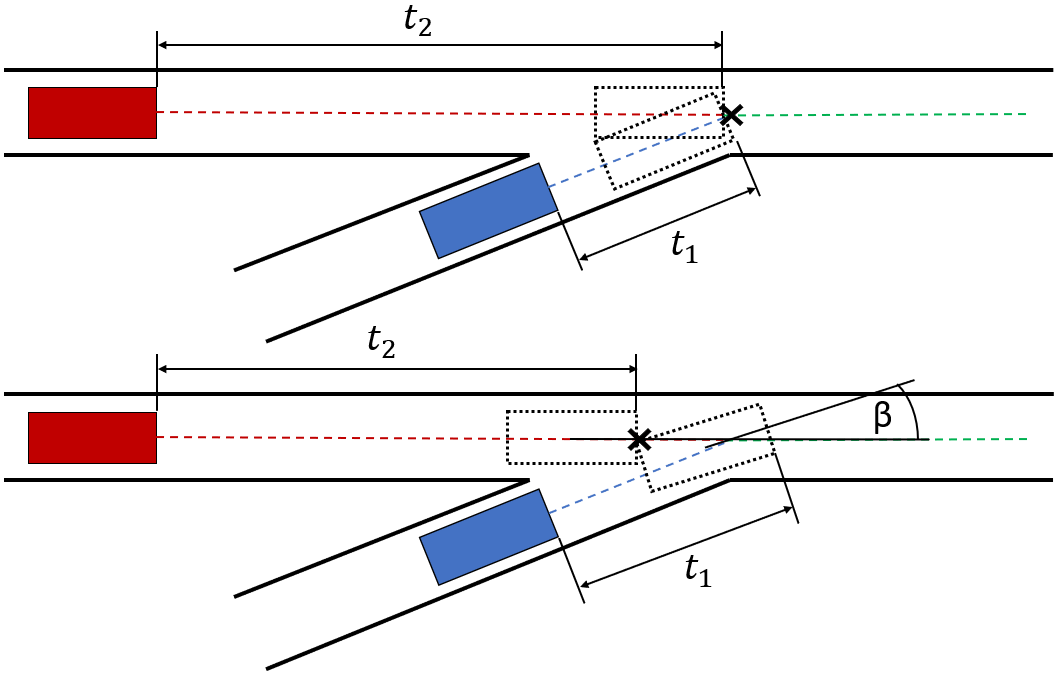}
	\caption{Original (upper)~\cite{INTERACTION_scheme} and modified (lower) calculation of $\Delta$TTCP conflict point (black cross) in order to not define merging points. Instead using a collision angle $\beta$.}
	\label{figure:dttcp_modification}
\end{figure}

Furthermore, the definition of conflict points is adopted for the $\Delta$TTCP. In simple cases like junctions, a conflict point (CP) is originally defined by the road layout. However, if the crossing point is not uniquely given by the road layout, as in the case of a ramp, the conflict point is derived from the trajectories driven. We adopt this concept by considering not only one but all possible conflict points. These are defined by the set of all points that both road users share on their real driven paths including their physical dimensions. Thereby, only the physical dimensions of the road users and not the infrastructure or situation are considered.
After this conflict point definition, the calculation is analog to the $\Delta$TTCP.
This modified Time to conflict point ($\Delta$mTTCP) is also mathematically similar to the $\Delta$TTCP.
Starting from the road users' current position, the Time to conflict point (TTCP) is determined as the time a road user takes to travel to a conflict point by assuming a constant speed and using the actual path driven.
In contrast to the $\Delta$TTCP, the time difference is determined not for one, but all conflict points, so that each vehicle pair has a set of time differences linked to individual conflict points. The smallest of those time differences is defined as the $\Delta$mTTCP (\ref{eq:dmTTCP}) and the concerning conflict point is defined as the critical conflict point (CCP).
\begin{align}
\Delta mTTCP = \min_{p \in CP}{|{mTTCP}^{p}_1 - {mTTCP}^{p}_2|} \label{eq:dmTTCP}
\end{align}
This more comprehensive and accurate definition of conflict points (see Fig.~\ref{figure:dttcp_modification}) requires neither the division of conflict points into static and dynamic nor detailed information about the infrastructure. So, it can be applied not only for vehicles but also for bicycles and pedestrians on sidewalks.
Calculating the TTCP for all possible conflict points causes a higher computational complexity compared to the original approach. However, to limit the complexity, a maximum prediction horizon of 5~seconds is used for trajectory prediction. As following scenarios are already covered by the THW, a minimum angle between the intersection of trajectories is utilized to filter them. For urban scenarios, it is set to 20~degrees because at that angle the mid-line of the rear vehicle would contact the side and not the rear of a car in front of it. Using that threshold merging scenarios or intersecting trajectories can still be detected. For highway scenarios, a much lower angle ($\beta_{min} =  2$ degrees) is used because of the parallelism of the vehicles and smaller yaw rates. This angle is derived from a potential lane change maneuver on a normal street within 100 meters.

With these adjustments, a variety of metrics can be applied robustly to any infrastructure. For a comprehensive view, four common metrics are used as detection types. These are chosen to supplement each other both in the type of scenario identification and its criticality assessment without overlapping much (Tab.~\ref{table:scope_metrics}). 
These metrics are calculated for all road user pairs and assigned to both respectively. Exceptions to this are THW and DRAC as these are each based primarily on the behavior of one road user. In these cases, the calculations are performed bidirectionally, so that both road users can have different DRACs and THWs.
In addition to these bilateral metrics, the WP is considered, too. It does not map the interaction between two road users, but registers forced stops. Those imply interactions with the traffic in general and therefore cover multiple possible interactions~\cite{INTERACTION_scheme}.

\begin{table}[h]
	\caption{Scope of relation indicators}
	\label{table:scope_metrics}
	\begin{center}
		\begin{tabular}{c|c|c|c|c}
			\textbf{Metric} & \textbf{Follow-up} & \textbf{Merging} & \textbf{Criticality} & \textbf{Coverage} \\
			\hline
			THW & ++ & - & o & o \\
			TTC & o & + & + & o \\
			DRAC & - & - & ++ & o \\
			$\Delta$mTTCP & - & ++ & o & + \\
			WP & - & + & o & ++ \\
		\end{tabular}
	\end{center}
\end{table}

\subsection{Individual Vehicle State}

Relation indicators are an important aspect, but can only capture part of a traffic scenario. By adding information about the individual driving states, the framework is extended. The observation is less relevant for interaction but more important for anomaly detection. For the description of the vehicle state, the bicycle model~\cite{bicycle_model} is chosen. So, just a few, but basic and characteristic parameters are observed (Tab.~\ref{table:vehiclestate_limits}). 
The most obvious value is the absolute speed. In addition, the different types of acceleration represent essential aspects of the vehicle maneuver. They do not only describe the change in the driving state, but also give a statement about forces and driving behavior. Additionally, the sideslip angle is considered as an extension of the force transmission and characterizes the contact between wheel and street. Finally, the yaw rate is added for a more precise characterization of cornering than just lateral acceleration. So, a set of five parameters is chosen to represent the vehicle state and influence the framework via detections. But since detections are only supposed to record interesting and unusual information, only exceedances of normal driving behavior are taken into account. These limits of normal driving maneuvers are based on results from publications~\cite{vehicle_limit_BEN15, vehicle_limit_DOU11}.

\begin{table}[h]
	\caption{Vehicle state detection types with thresholds}
	\label{table:vehiclestate_limits}
	\begin{center}
		\begin{tabular}{c|c|c}
			\textbf{Parameter} & \textbf{Threshold} & \textbf{Range [$km/h$]}\\
			\hline
			Velocity & see chap.~\ref{sec:detections_infrastructure} & -\\[0.5ex]
			Lon Acceleration & $4 m/s^2$ & $v \leq 50$ \\[0.5ex]
			& $4m/s^2 - 2m/s^2 \frac{v - 50km/h}{100km/h}$ & $50 < v \leq 100$ \\[0.5ex]
			& $2m/s^2$ & $ 100 < v$\\[0.5ex]
			Lat Acceleration & $2.5m/s^2 + 4.5m/s^2 \frac{v}{40km/h}$ & $ v \leq 40$ \\[0.5ex]
			& $7m/s^2$ & $40 < v \leq 50$ \\[0.5ex]
			& $7m/s^2 - 4m/s^2 \frac{v-50 km/h}{50 km/h}$ & $50 < v \leq 100$ \\[0.5ex]
			&$3m/s^2$ & $v < 100$\\[0.5ex]
			Yaw rate & $\frac{50}{180} \pi \deg/\sec $ & $ v \leq 50$ \\[0.5ex]
			& $\frac{15}{180} \pi \deg$ / $\sec$ & $50 < v$\\[0.5ex]
			Sideslip angle & $10 \deg$ & - \\
		\end{tabular}
	\end{center}
\end{table}

\subsection{Context-Related Behavior}
\label{sec:detections_infrastructure}

\begin{figure}[thpb]
	\centering
	\includegraphics[width=\linewidth]{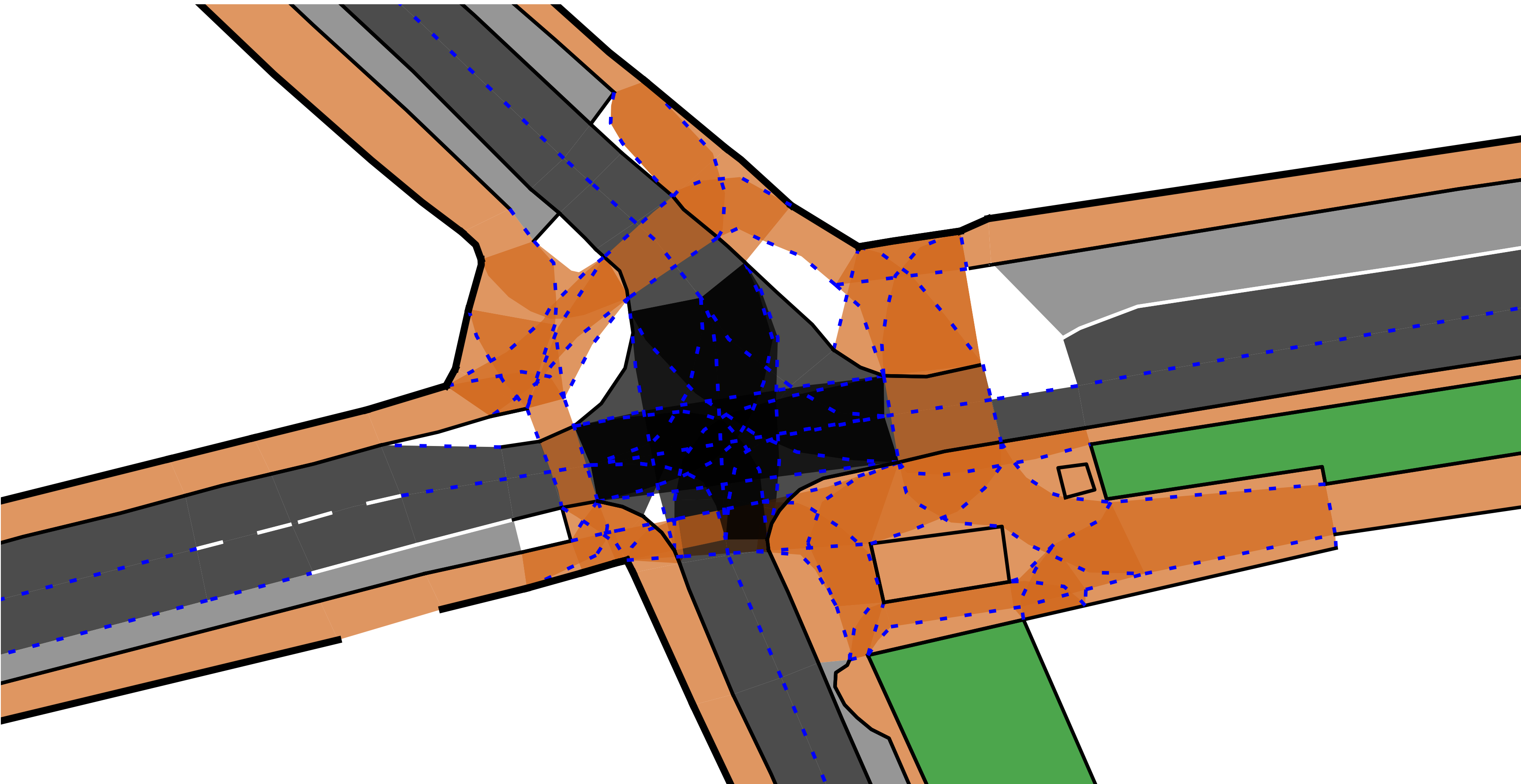}
	\caption{Digital map of the intersection \textit{Frankenberg}~\cite{inD} with colored region types as provided in the dataset (black: street; gray: parking lot; orange: walkway; green: grass verge) and intersecting regions (darker than normal colors)}
	\label{figure:inD_frankenberg_map}
\end{figure}

While both previous groups of detection types do not consider traffic rules or guidance, the latter is examined in more detail below. For this purpose, road user behavior is subdivided into two categories: the first with respect to traffic rules and the second to other road users. Therefore, the peculiarity of the infrastructure of the dataset is used. So, it must be divided into logical driving \textit{regions} (see Fig.~\ref{figure:inD_frankenberg_map}) as set out in \ref{sec:data_requirements}. The regions represent possible traffic routes or sections of those. For example, a two-lane street is described at least with two regions. A t-junction is described by at least six regions because of the six possible routes. On an intersection, where several driving routes and walkways cross, an overlapping of regions is also possible.

In order not to be bound to national peculiarities when checking traffic rules, they are kept general. For this reason, only three rules are taken into account referring to the dataset requirements:
\begin{itemize}
	\item Local speed limit
	\item Designated area usage
	\item Driving direction
\end{itemize}
These rules are evaluated for all vehicles in their respective regions. If a road user can be assigned to more than one region, it is assumed that the road user behaves correctly, if he behaves correctly in at least one of the assigned regions. This is particularly relevant when considering the direction of travel, for example, if regions overlap in the center of an intersection.

In addition to the rules given by the infrastructure, the behavior of vehicles, pedestrians and other vulnerable road users are examined in relation to the norm. This norm may differ from infrastructure rules due to common rule violations e.g. if a designated lane is blocked. For this purpose, driving behaviors are compared on a local level as a combination of speed ($v$) and orientation ($\psi$). They are collected in each region and grouped by using a DBSCAN clustering approach~\cite{DBSCAN}. Therefore, the distance ($D$) between two data points ($i,j$) is defined using a Mahalanobis norm. \begin{align}
	{D}_{i,j} = \sqrt{(\psi_1 - \psi_2)^2 + \Bigl(\frac{v_1 - v_2}{\max\{\max\{v_1, v_2\}, 1.5\}}\Bigl)^2} \label{eq:adaptive_mahalanobis}
\end{align}
The norm is scaled with respect to the velocity, since the usual deviation of the direction of movement decreases with increasing travel speed. So, pedestrians and vehicles on highways can be discussed equally.
Another advantage is, that the approach can be used for sidewalks as well as for freeways.

A detection is only found if a driving state cannot be assigned to a cluster or if the cluster of the state makes up less than 10 percent of the total quantity. Applying this rule, unusual driving maneuvers can be detected with reference to the infrastructure without having a look at usual behavior a priori. So, driving against the direction of travel can be regarded as normal if a significant amount of objects do so. For example, this common misbehavior can be caused by a blockade of the own lane. Since it is might still be an interesting situation, it is detected by infrastructure rules. Nevertheless, it is rated less important because it is only assigned to the infrastructure rules, but not to a specific cluster.
To examine the clustering itself, values for $\epsilon$ ($= 0.7$) and \textit{min\textunderscore samples} (\eqref{eq:min_samples}) are chosen by experiment. Thereby \textit{min\textunderscore samples} is set adaptively with regards to the number of detections ($|M_\text{input}|$) to take care of the wide span of possible input data.
\begin{align}
min\_samples = 2 + 0.01 \cdot |M_\text{input}| \label{eq:min_samples}
\end{align}

On a more abstract level, a similar procedure is used. Not the driving states, but whole trajectories are grouped by the HDBSCAN algorithm~\cite{HDBSCAN}. Therefore, the Fr\'{e}chet norm~\cite{frechet_norm} is used to determine the distance. Just like the previous clustering, the number of trajectories ($|M_\text{Trajectories}|$) can vary. So, the \textit{min\textunderscore cluster\textunderscore size} is set adaptively as well.
\begin{align}
	min\_cluster\_size = 1 + 0.005 \cdot |M_\text{Trajectories}| \label{eq:min_cluster_size}
\end{align}

\begin{figure}[thpb]
	\centering
	\includegraphics[width=\linewidth]{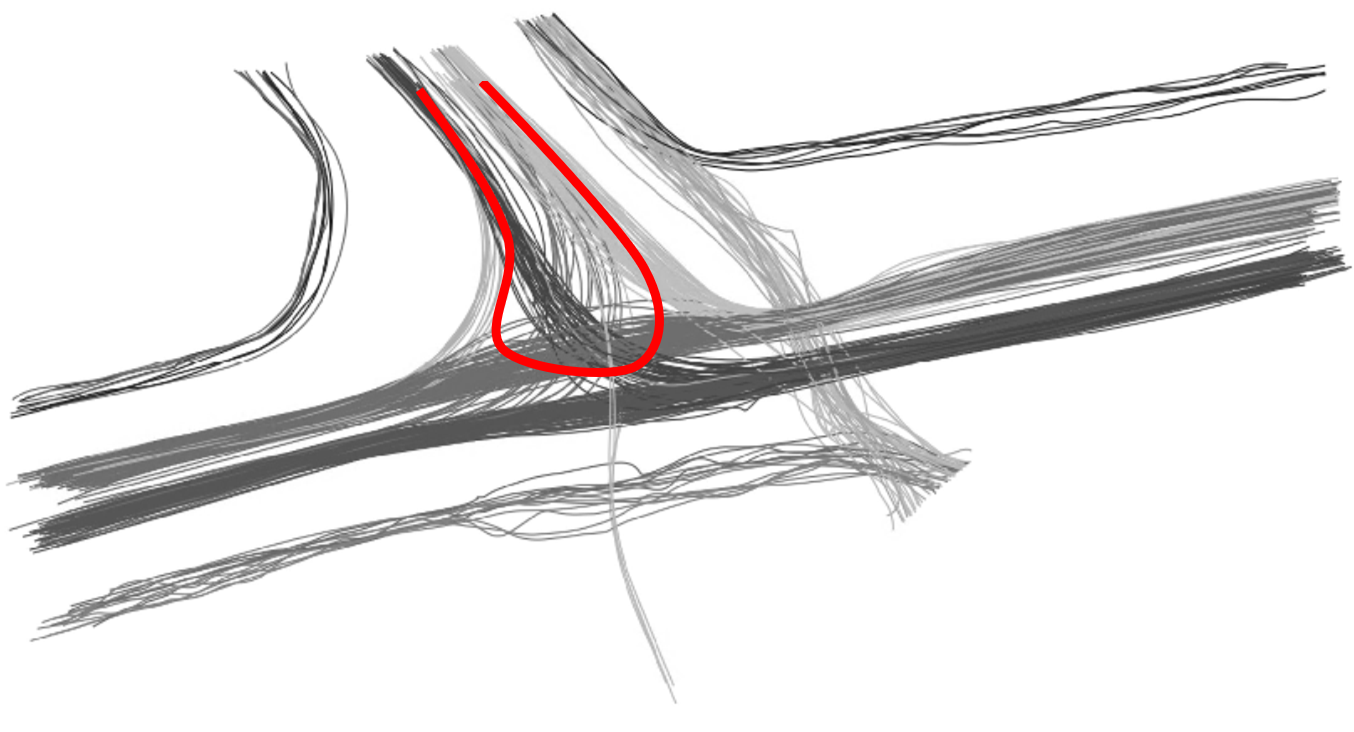}
	\caption{ trajectories (different grays) at an intersection with an additional anomaly (red)}
	\label{figure:anomaly_in_trajectory_clustering}
\end{figure}

Those points that do not belong to any significant cluster are outliers and declared as detections (see Fig.~\ref{figure:anomaly_in_trajectory_clustering}). For further classifying outliers the possibility that even small deviations can cause a trajectory to be an outlier is taken into account. Therefore, the value of the Fr\'{e}chet distance to the nearest trajectory within a cluster is used as a measure of anomaly.

\subsection{Comparability of Detections}
\label{sec:comparability_of_detections}
		
The fourteen presented detection types (see \ref{sec:detections_interaction}~-~\ref{sec:detections_infrastructure}, Tab.~\ref{table:detection_scoring}) out of the three areas are combined in chapter~\ref{chap:combining_scores} to create interaction, anomaly and relevance scores. For this purpose, each detection is scored according to its characteristics. To create a relative weighting between the detection types, $THW(1s) = 1$ is set as a reference score. Based on this metric, the other components are scaled relative to the THW with regards to their criticality and potential interestingness (Tab.~\ref{table:detection_scoring}). Thereby, binary detections are assigned as fixed values. Correspondingly, real-valued detections are scored as those. To not distort the result, an upper limit is set so that errors in the input data do not dominate. Of course, the scorings tend to be subjective, but they are weighted by simple rules: As the detections refer to the THW scoring, detections of different intensities are evaluated similarly. Nevertheless, situations with high conflict potential are evaluated higher, because they combine both: low and high-intensity detections. In order to give the user the possibility to prefer either the occurrence or the intensity of interaction, the criticality factor $\kappa$ is proposed. This factor is set to 1 by default. An increase leads to a focus on intensive interactions.
Compared to these relation indicators mainly anomaly-based detections are weighted higher, so that anomalies stand out from ordinary interactions. The vehicle state and behavior detection types themselves are weighted equally so that none of them dominates the score.

\begin{table}[h]
	\caption{Scoring of detection types}
	\label{table:detection_scoring}
	\begin{center}
		\begin{tabular}{c|c|c}
			\textbf{Detection type} & \textbf{Scoring} & \textbf{Max} \\
			\hline
			THW 				& $\frac{1}{THW}$		& 2	 \\
			$\Delta$mTTCP 		& $\frac{1}{\Delta mTTCP} \frac{4}{{mTTCP}_1 + {mTTCP}_2}$		& 4	 \\
			TTC 				& $\frac{2 \kappa}{Value}$		& 2 $\kappa$	 \\
			DRAC 				& $\frac{\kappa}{5} \cdot Value $		& 2 $\kappa$	 \\
			WP			 		& $\sqrt{Value}$		& 7.75	 \\
			Lon acceleration	& $0.1 \cdot |Value - Limit|$		& 10	 \\
			Lat acceleration	& $2 \cdot |Value - Limit|	$	& 20	 \\
			Sideslip angle		& $25 \cdot |Value - Limit|$		& 8.725	 \\
			Yaw rate	 		& $|Value - Limit|$		& 3.141	 \\
			Area usage 	& $5$		& -	 \\
			Driving direction 	& $4$		& -	 \\
			Velocity	 	& $\frac{10}{Limit} (Value - Limit)$		& 10	 \\
			Driving behavior (\ref{sec:detections_infrastructure})	& $1.2 \cdot Value$		& -	 \\
			Trajectory (\ref{sec:detections_infrastructure})		 	& $Value	$	& 10	 \\
		\end{tabular}
	\end{center}
\end{table}

\section{COMBINING DETECTIONS TO ABSTRACT SCORES}
\label{chap:combining_scores}

As described, the interaction-, anomaly- and relevance scores are based on the introduced detections which can be calculated for each timestep. Within the scope of this paper, each recording was considered individually. A larger set of trajectories like the combination of several recordings within an infrastructure would be also possible to use as an input for the framework. However, due to changing traffic conditions like congestion or time depending behavior, temporally different recordings may only be less meaningful because of the change in boundary conditions. 
Based on the determined detections, the scores are calculated for every traffic participant at any given time in a first step (chap.~\ref{sec:interaction_score} - \ref{sec:relevance_score}). A hierarchical combination of these scores is proposed in chap.~\ref{sec:hierachical_kombination}.

\subsection{Interaction Score}
\label{sec:interaction_score}

\begin{figure}[thpb]
	\centering
	\includegraphics[width=\linewidth]{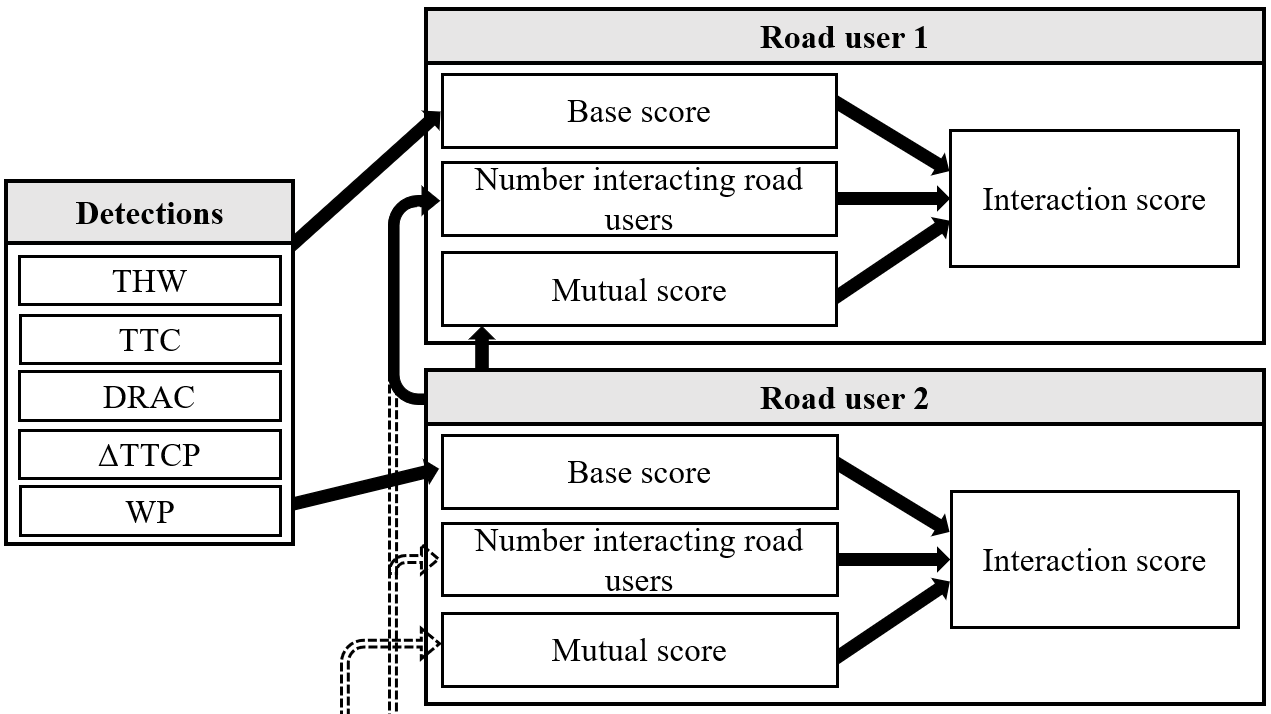}
	\caption{Flowchart of the interaction score calculation. Beginning with the road user related detections, three influences are calculated and summed up for interaction score.}
	\label{fig:flowchart_interaction_score}
\end{figure}

A core requirement for trajectory datasets is high interactivity between traffic participants. This interplay is represented by the interaction score (see Fig.~\ref{fig:flowchart_interaction_score}). For calculating it, the vehicle relation indicators (Chap.~\ref{sec:detections_interaction}) are used as its base: all detections belonging to the relation indicators of a traffic participant are scored (see Tab.~\ref{table:detection_scoring}) and summed up to a base score. 
\begin{align}
	{S}_\text{Base} &= \sum_i^\text{Detections}{S}_{i} \label{eq:interaction_base_score}
\end{align}
Since the base score is limited to the interaction between two road users, two further influences are considered: the freedom of movement of the observed vehicle and the opportunity of participating road users to handle a specific situation. 

If a road user interacts with multiple other traffic participants simultaneously, multiple detections are assigned to the road user at the same point in time. Handling such scenarios is more difficult since the road user has to consider the behavior of different traffic participants. Therefore, the higher the number of different interacting partners in the detections of the observed vehicle ($R_\text{Observed}$) is, the higher the interaction is rated. 
Just as relevant as the freedom of movement of the observed vehicle are the numbers of interacting vehicles of the partners themselves ($R_i$). For example, if the observed vehicle wants to drive onto a freeway, but the lane is blocked by another vehicle, the interaction could be defused by a lane change of this second vehicle. However, if the second road user itself interacts with other participants who are preventing a lane change, the situation becomes more interactive for the observed vehicle, too. Therefore, this mutual interaction (${S}_\text{Mutual}$) is considered separately (\ref{eq:interaction_mutual}).
After calculating these three aspects, they are combined to the punctual interaction score of a road user (\ref{eq:interaction_overall}).
\begin{align}
    {S}_\text{Mutual} &= \sum_{i}^{\substack{\text{Interacting} \\ \text{vehicles}}} \sum_{j}^{\substack{\text{Mutual} \\ \text{detections}}} 0.1 \cdot R_i \cdot S_{j} \label{eq:interaction_mutual} \\
	{S}_\text{Interaction, Punctual} &= {S}_\text{Base}\cdot(1 + 0.1 \cdot{R}_\text{Observed}) + {S}_\text{Mutual} \label{eq:interaction_overall} 
\end{align}

\subsection{Anomaly Score}

\begin{figure}[thpb]
	\centering
	\includegraphics[width=\linewidth]{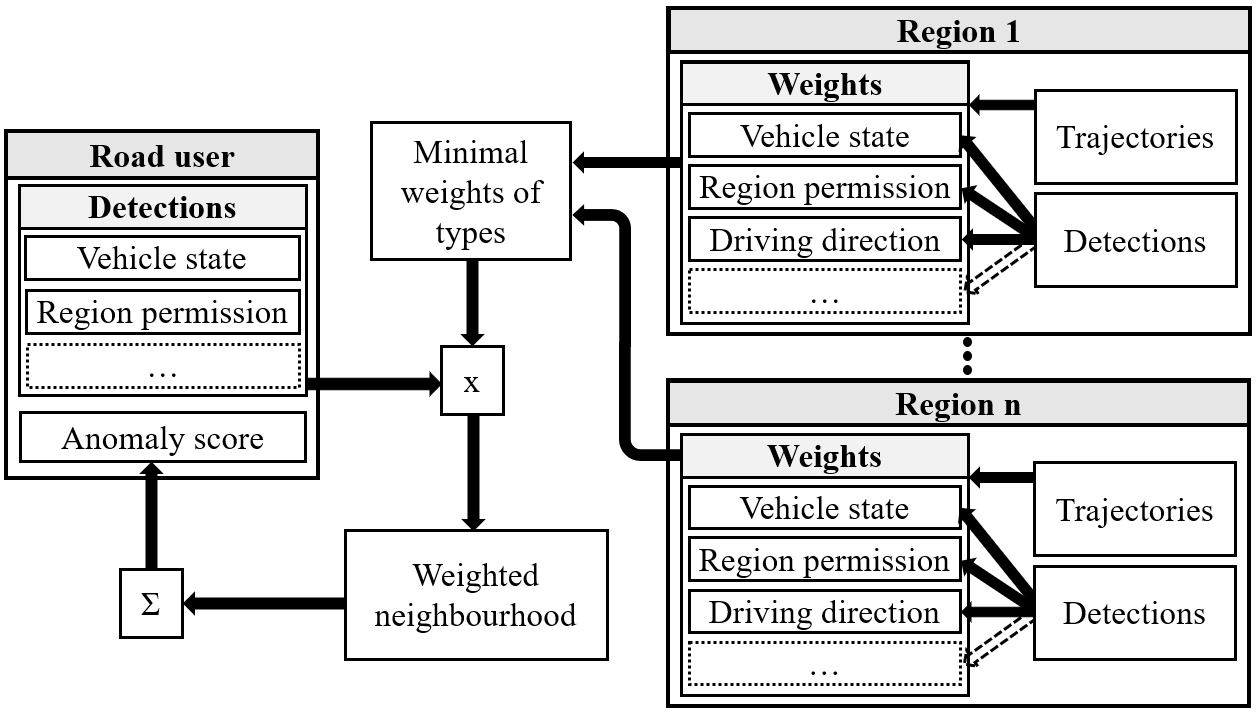}
	\caption{Flowchart of the anomaly score calculations. Each detection is weighted by a context provided by the belonging regions and all of them are summed up afterwards. If the affiliation is unclear, additionally, the neighbourhood is taken into account.}
	\label{figure:flowchart_anomaly_score}
\end{figure}

Since the framework follows the goal to describe datasets comprehensively, a sole consideration of interaction is not sufficient. Often, it is the interest to find unusual and interesting situations~\cite{driving_factors}. Therefore, the anomaly score focuses on the rarity of events. In order to follow this principle each detection type is weighted depending on its rarity within the context ($\gamma_i$).
To correctly interpret anomalies even in larger infrastructures and not to bundle different phenomena, an individual context is defined for each region.
Within these regions the number of all detections of a detection type belonging to a region ($M_\text{Type,Region}$) is considered as well as the number of road users passing the same region ($U_\text{Region}$). Those attributes form the weights of the contexts. By using the weights, the detections can be scaled before they are summed up to the anomaly score (see Fig.~\ref{figure:flowchart_anomaly_score}). 
In accordance with the rarity of anomalies, the amount of detections is considered twice: 
Firstly, to normalize the number of detections, so that a higher amount does not directly lead to a higher score because that would lead the anomaly concept ad absurdum.
Secondly, the framework follows the approach, that a rarity is worth more, if the quantity of road users is higher. To limit this effect, the impact of the amount of anomalies is diminished by the root.
\begin{align}
	\gamma_i &= \frac{1}{M_{\text{Type,Region}, i}} \frac{U_{\text{Region}, i}}{{M_{\text{Type,Region}, i}}^{\frac{1}{2}}} \\ 
	{S}_\text{Anomaly, Punctual} &= \sum_i^\text{Detections} {S}_{i} \cdot \gamma_{i} \label{eq:anomaly_overall}%\\
	%&= \sum_i^\text{Detections} {S}_{i} \cdot \frac{|U_\text{Region}|}{|M_\text{Detectiontype,Region}|^{\frac{3}{2}}}
\end{align}

In contrast to the interaction score, anomaly is not limited to a few detection types, but considers all of them, to ensure a comprehensive view of the dataset. 
For example, waiting (waiting period) before a crosswalk is common, but a stationary car on a freeway with fast-moving traffic is unusual. 
Since the context is set up for each region, it is possible to differentiate standing on a highway from stopping in front of a crosswalk and tracking those anomalies on both trajectory and dataset level, too.
However, regions may overlap and vehicles may belong to several regions at the same time. In the case of overlapping regions, it is assumed that a road user behaves normally. So, the least unusual context is selected for calculating the detection score. Unluckily, the problem of the affiliation of a road user to a region and the correct context for a detection is more complex since both, road user and region each cover a certain area and can overlap to different degrees. A vehicle is assigned to a region as soon as it overlaps with the region with at least two square meters or occupies more than 50 percent of the region. If this results in a road user being assigned to more than one region, the contexts of those regions are taken into account and the detection is weighted by as many regions as needed to assign 100 percent of the vehicle to regions. Thereby, the regions used for the calculation are chosen depending on their weights to minimize the overall anomaly. To finally calculate the score the detection region weights are proportionally taken into account with regards to the intersecting area between road user and region. This method is used to distinguish vehicles that only touch a sidewalk with their bumper from those that are completely on it.
The same concept applies to vulnerable road users (VRU). However, these are usually only given as points and like to stretch existing rules e.g. when walking on the street directly next to a sidewalk. To take these points into account, an area is defined around the VRU as a circle with a radius of 2.5 meters. It is used for the described calculation as the area vehicles have by default. It seems a lot, but considering, that 2 square meters are enough to fully assign a road user to a region and the least unusual region is chosen, the method reduces the misdetections due to inaccuracy and minor rule violations by VRUs.

\subsection{Relevance Score}
\label{sec:relevance_score}

If interaction and anomaly are considered separately, only limited statements can be made about the relevance of traffic scenarios. For example, abnormal behavior without interacting with a second traffic participant normally has little relevance e.g. for validation. Similarly, an anomaly makes an interactive situation more interesting. To account for this, the relevance score ($S_\text{Relevance}$) is introduced. It closes the gap resulting from the two previous scores. It is not based on the detections directly, but it is defined by the combination of the punctual interaction ($S_\text{I}$) and anomaly score ($S_\text{A}$). 
\begin{align}
{S}_\text{Relevance, Punctual} = {S}_\text{I} \cdot {S}_\text{A} + \gamma_\text{I} \cdot {S}_\text{I} + \gamma_\text{A} \cdot {S}_\text{A}
\label{eq:relevance_score}
\end{align}

In order to consider highly interactive scenarios as well as anomalous in the evaluation framework, they are included in the score on a subordinate scale. Thereby, the different scoring of the individual detections in favor of the anomaly-related detection types is compensated through weights ($\gamma_{I} = 5$, $\gamma_{A} = 0.1$). The weights are chosen, so that default maximum values of interaction and anomaly have a similar impact. Depending on the application, other weightings can be quite conceivable. 

\subsection{Hierachical Combination of Scores}
\label{sec:hierachical_kombination}

Defined scores were previously determined for timesteps of individual road users. This is not enough for a comprehensive overview, because a few hundred thousand scores can be calculated for a normal recording over 15 minutes. Therefore, these scores are converted into more abstract layers. Starting with the defined scores within one timestep, the scores are determined for trajectories, local regions and finally for whole recordings. Therefore, two different methods are used: One is purely detection-based and the other one only dependent on abstract scores. Interaction and relevance scores have to be considered differently for more abstract scores because they are additionally based on other influences like the number of interacting road users. In order to generate trajectory scores from the punctual scores, only the time series of the punctual score itself is used to derive the trajectory score. Only the positive change of the series is summed up. 
\begin{align}
	{S}_\text{g, Trajectory} &= \int \max\Biggl(\frac{\text{d}{S}_\text{g, Punctual}}{dt}, 0\Biggl) dt \label{eq:positive_summed_up} \\
	g &\in [\text{Interaction}, \text{Relevance}]
\end{align}
This method is preferred to ordinary integration, since an unchanged state is usually not very interesting, regardless of the duration. For example, a continuous distance and velocity between two vehicles are not relevant. However, this scenario would become interesting when the situation changes, e.g. because the vehicle in front brakes and expects a reaction of the following vehicle. 
The same approach can be used analogously for regions by summing up the scores of all trajectory parts that are assigned to the region. Additionally, due to the previously determined layers, a dataset score is defined as the sum of the individual trajectories, which characterize the dataset extensively.
\begin{align}
	{S}_\text{g, Dataset} = \sum_j^\text{Trajectories} S_{g, \text{Trajectory}, j} \label{eq:summed_up_dataset}
\end{align}

A disadvantage of this method is the elimination of simultaneously occurring opposite effects: If a vehicle increases the distance to a vehicle in front, the THW increase and so the detection score itself decreases. But if the conflict with a third vehicle grows, this second detection score increases. Due to the punctual interaction score definition, both detection scores are summed up and so the changes partially balance each other out in $S_\text{Punctual}$. As a result of this, the trajectory score is underestimated. This issue does not have a big impact, but since the anomaly score is based directly on the detections, another approach can be used. For the creation of the abstract scores ($S_\text{Anomaly}$), the detections within the observed scope are collected and the maximum value of the individual detections are summed up. So all detections can be considered in their maximum intensity. In order to take a temporal extension into account, the detections are evaluated for each region. So it becomes more important if a detection occurs for a longer time.
\begin{align}
	{S}_\text{Anomaly, i, j} =\sum_{i}^{\substack{\text{   } \\ \text{Regions}}} \sum_{j}^{\substack{\text{  Road  } \\ \text{users}}} \sum_{k}^{\substack{\text{Detection} \\ \text{types}}} \max{S}_{i,j,k} \label{eq:combination_anomaly}
\end{align}

\section{EVALUATION}

In order to check the suitability of the framework, three different approaches are used. In a first step, a study is executed to check the correspondence between human perception and the framework. 
Although a study to check human perception of relevance would be best, this is not easily achievable. Reasons for this are that both, that anomaly and as a result of that relevance as well take the context of recordings into account. Therefore, study participants would have to overlook a large part of the record to rate specific situations. However, this long preparation time would lead to fatigue of the participants and thus to inaccurate results.
Therefore, we only test the human perception of interaction as a cornerstone of the framework. 
In contrast to the other scores, it offers the possibility of intuitively evaluating individual situations without further knowledge about the whole dataset. So, situations across multiple infrastructures can be checked without exhausting the participants too much. 
In a second step, this score is compared with existing approaches. That analysis is distinct from the study since there is no framework analyzing situations similar extensively, but focusing on bilateral interactions.
In a third step, anomaly and relevance are analyzed by applying these scores in different levels of abstraction to three actual datasets in urban scenarios. In this context, existing datasets are analyzed and compared with each other. That approach is used because the scores can not be tested against previous schemes. Anomaly has not yet been considered in such a holistic approach and relevance not at all in the literature. The evaluation is limited to the datasets inD \cite{inD}, rounD \cite{rounD} and INTERACTION \cite{INTERACTION_dataset} because they belong to few actual datasets fulfilling the requirements and represents a broad range of scenarios.

\subsection{Interaction Score Study}
\label{sec:interaction_score_study}

\begin{figure}[thpb]
	\centering
	\subfloat[Intersection \textit{Frank- enberg}~\cite{inD}]{
		\includegraphics[width=0.29\linewidth]{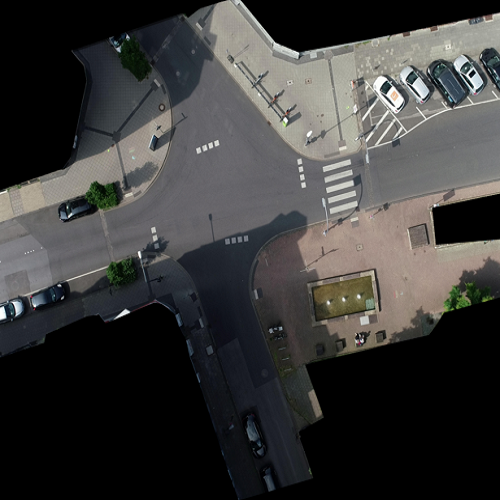}
		\label{fig:infrastructure_frankenberg}
	}
	\hspace{0.005\linewidth}
	\subfloat[Intersection \textit{Aseag} \cite{inD}]{
		\includegraphics[width=0.29\linewidth]{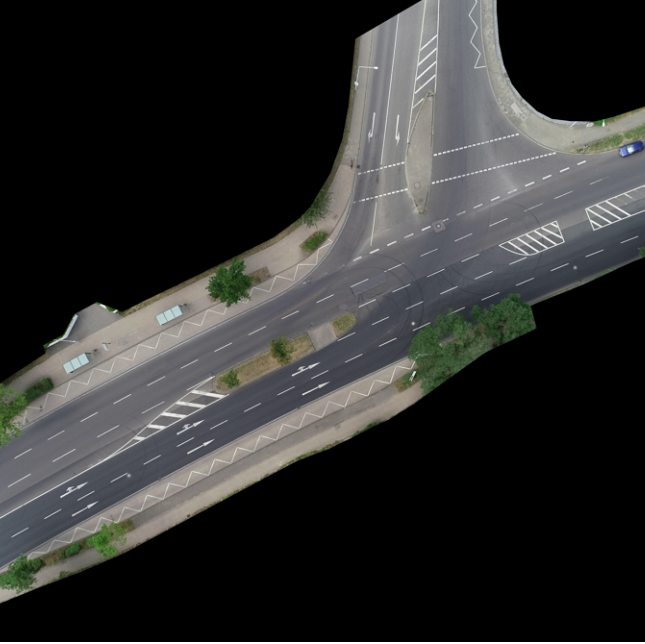}
		\label{fig:infrastructure_aseag}
	}
	\hspace{0.005\linewidth}
	\subfloat[Roundabout \textit{Neu- weiler}~\cite{rounD}]{
		\includegraphics[width=0.29\linewidth]{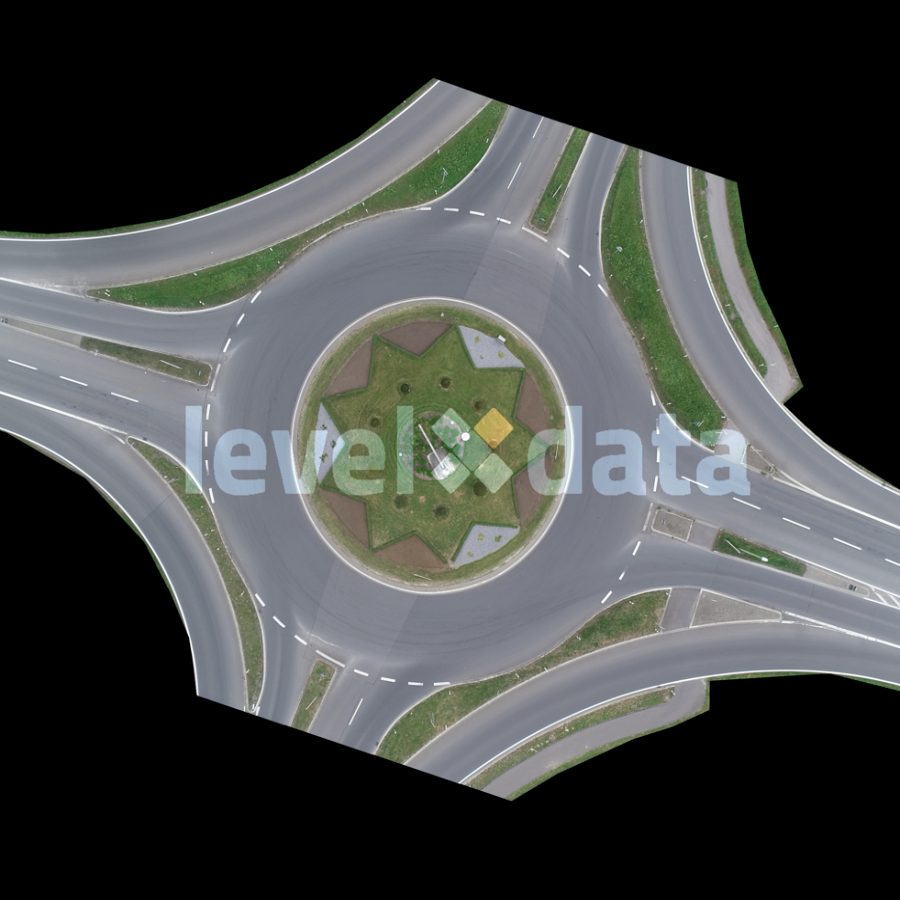}
		\label{fig:infrastructure_neuweiler}
	}
	\caption{Bird-eye perspective of observed infrastructures of the study.}
	\label{fig:survey_infrastructure}
\end{figure}

In order to check the correspondence between human perception of interaction and the interaction score, a study is carried out. For this purpose, twenty-four persons with a background in automated driving and automotive engineering estimate the interaction in given situations from the perspective of a selected road user. Therefore, they are provided with twenty aerial images, each for a single traffic situation. The participants score the situations subjectively and with regards to their experience on a scale of 0 - 10. These twenty situations belong each to one of three recording sites from two different datasets (see Fig.~\ref{fig:survey_infrastructure}). As these recording sites include a roundabout, a t-crossing and an intersection with a zebra crossing a wide range of scenarios are considered. For the selection of the situations, the number of participating road users, vehicle constellations and their velocities are varied. To enable the participants to rate the situations adequately, a short video of the traffic recorded in the dataset is shown to get them used to the bird-eye's perspective and the common traffic flow. Thereby, no reference score is given to get the uninfluenced interaction impression of the subjects.

\begin{figure*}[t]
	\centering
	\subfloat[Overall average]{
		\includegraphics[width=0.25\linewidth]{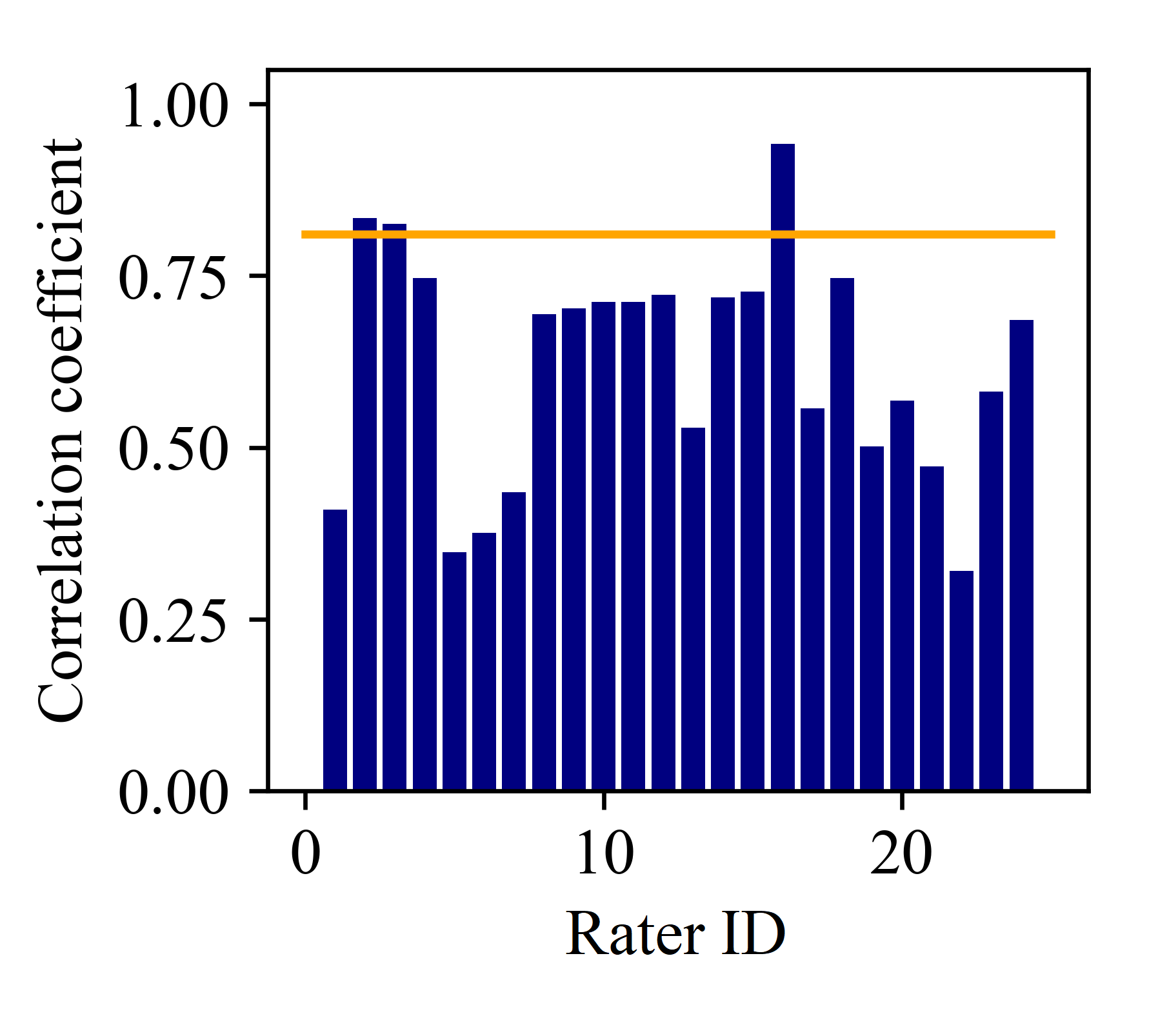}
		\label{fig:pearson_average}
	}
	\subfloat[Intersection \textit{Frankenberg}]{
		\includegraphics[width=0.237\linewidth]{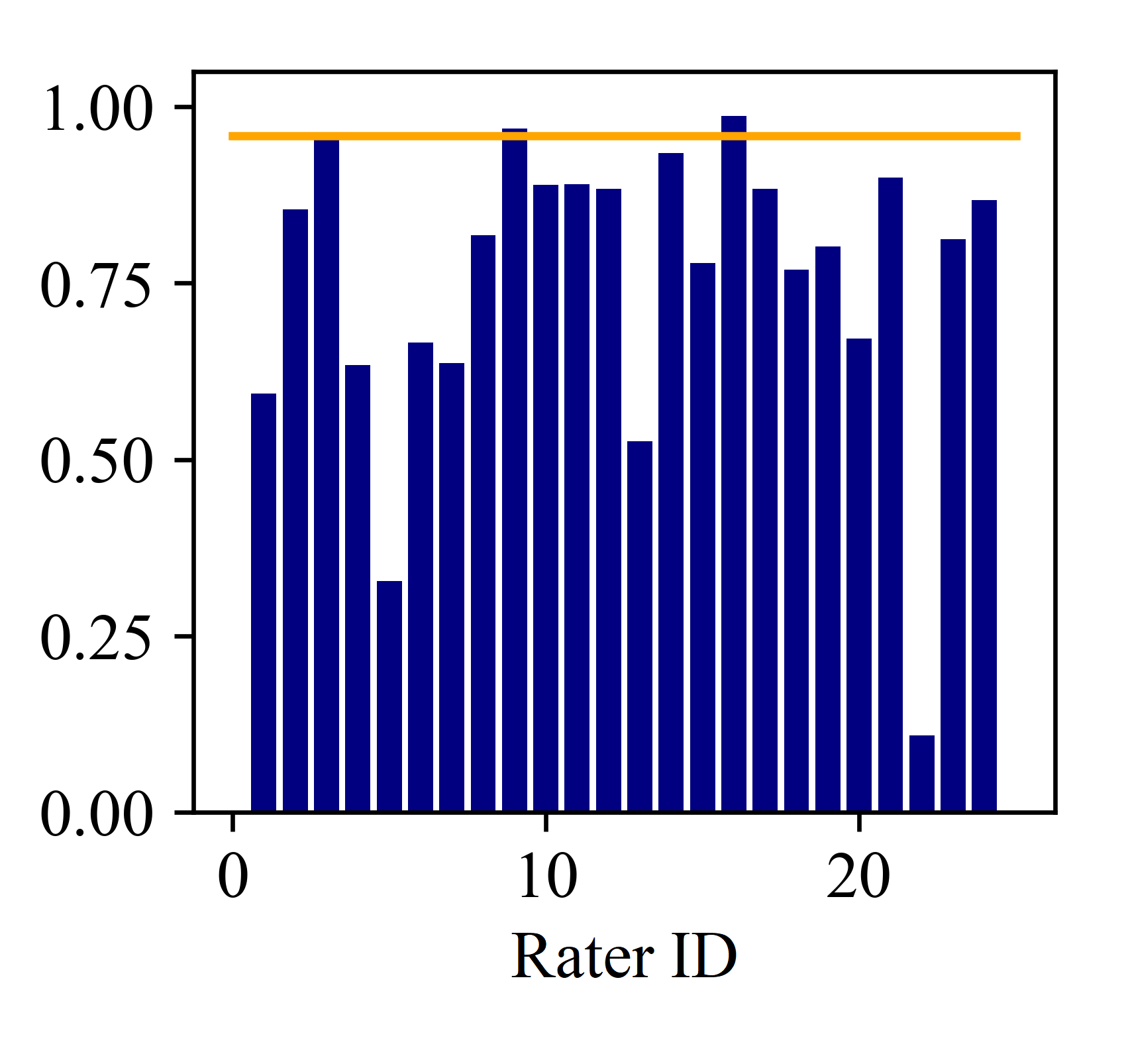}
		\label{fig:pearson_frankenberg}
	}
	\subfloat[Roundabout \textit{Aseag}]{
		\includegraphics[width=0.237\linewidth]{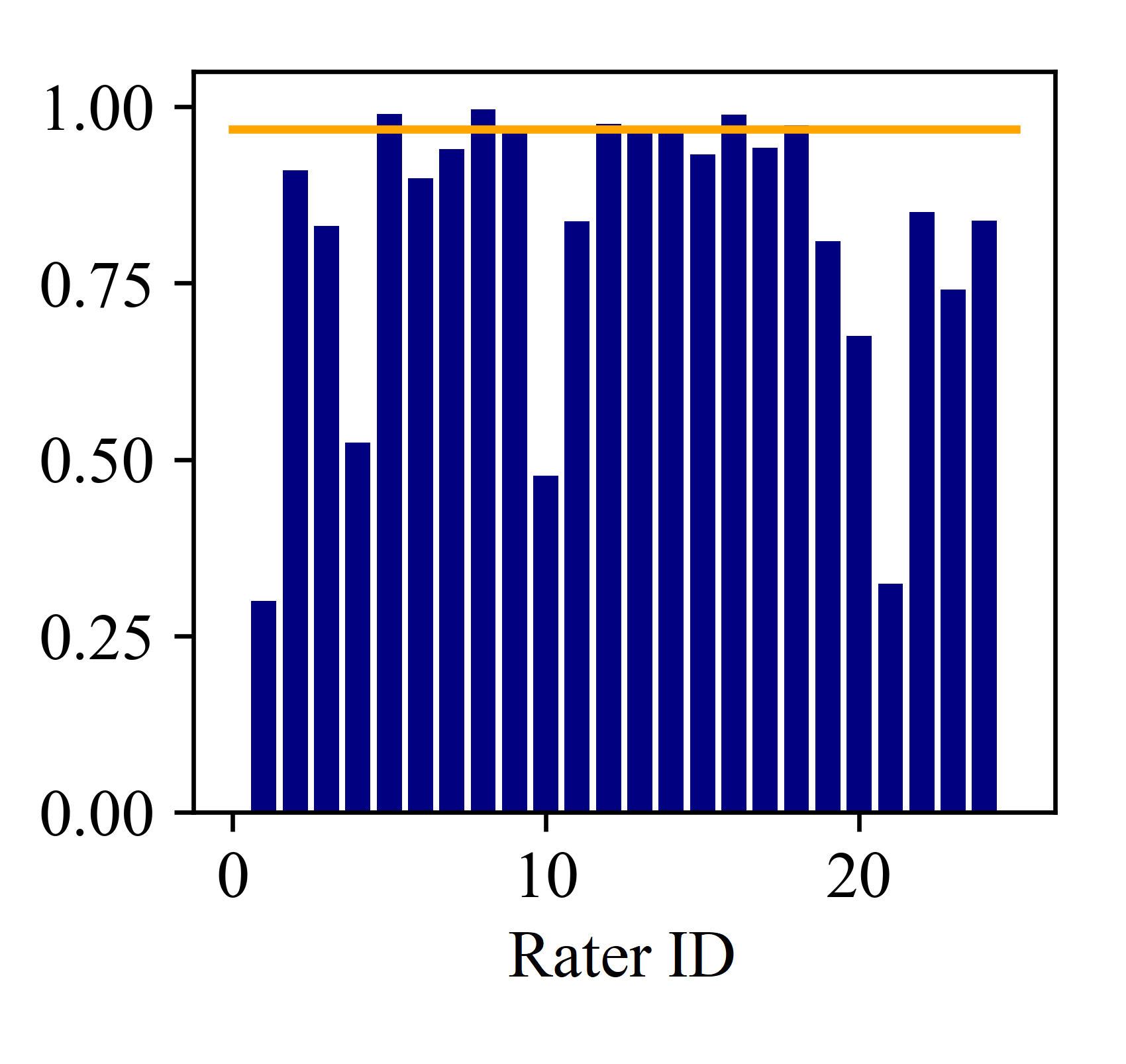}
		\label{fig:pearson_aseag}
	}
	\subfloat[Roundabout \textit{Neuweiler}]{
		\includegraphics[width=0.237\linewidth]{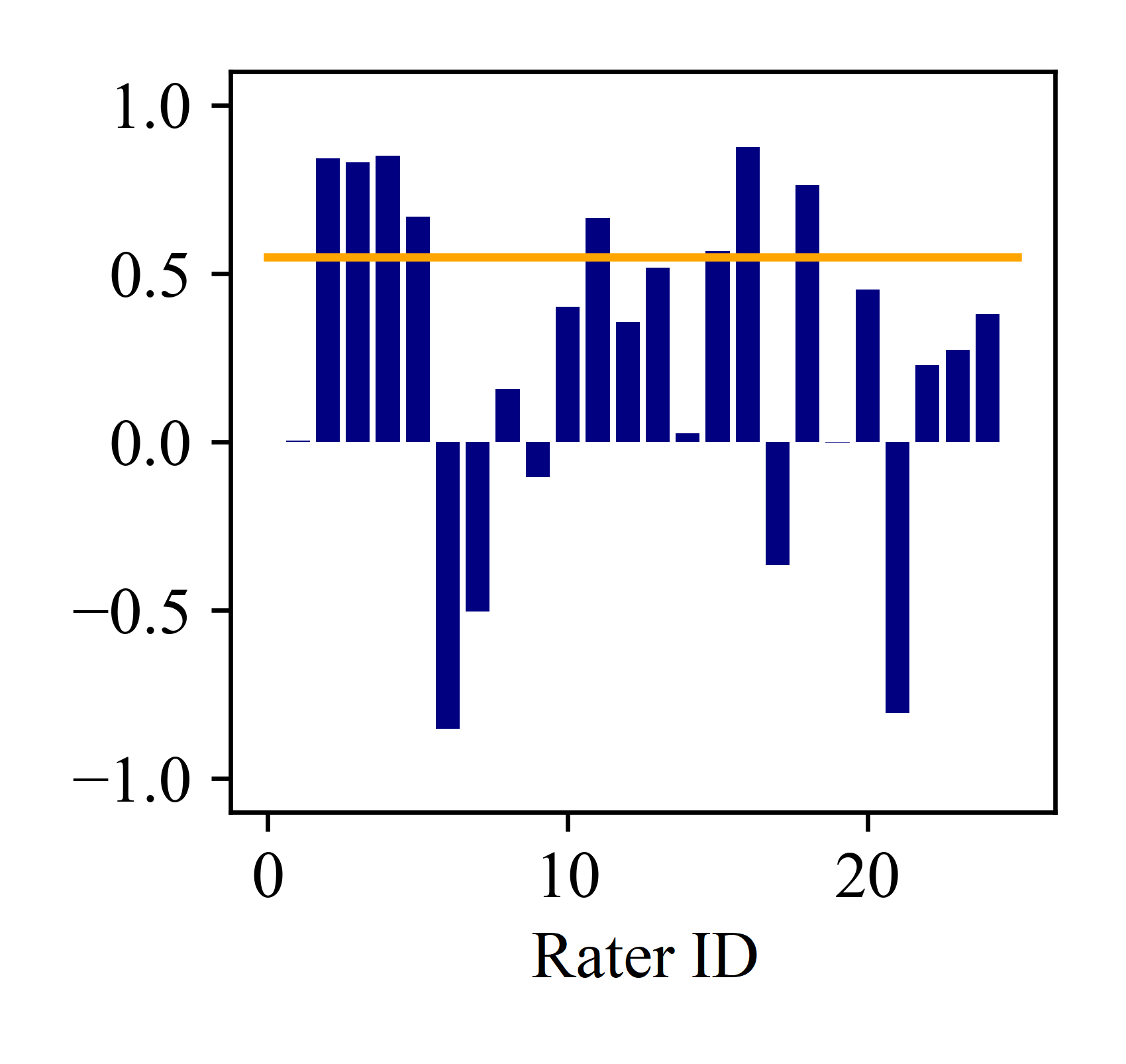}
		\label{fig:pearson_neuweiler}
	}
	\caption{Correlation coefficient (Pearson r) of each rater values compared to the calculated interaction scores of each rater (blue bars) and the average value of raters (orange line)}
	\label{fig:pearson_overview}
\end{figure*}

Because of the fixed range for the participants' interaction ratings, the interaction score is scaled to this range for the study. Despite a wide span (average 6.7) of the volunteer's ratings within the individual situations, a high correlation between the interaction score and the ratings can be observed for most of the volunteers (see Fig.~\ref{fig:pearson_overview}). Comparing these with a baseline correlation between the number of interacting road users and the human perception of interaction ($r = 0.34$), the interaction score fits the perception significantly better. Thereby, interacting road users are those that are either near to the observed road user or that intersects the road users future path within a reasonable time. Only at the roundabout \textit{Neuweiler} the perception of few participants deviates from the score, but is in average still significantly higher than the correlation to the number of interacting vehicles. However, if the ratings are averaged over the participants, the correlation is noticeably higher. That indicates that the framework reflects the average perception of the subjects better than the opinion of an individual expert. The average is used instead of e.g. the median because the perception of interaction tends to be subjective and the opinion of all participants should be included.
When comparing the absolute values, a similar picture emerges. While the scores normally differ by an average of 2 points between participant and framework, the average score is only 0.8 points higher than the interaction scores. It is noticeable that especially following maneuvers and conflicts with large prediction periods are evaluated differently. Classical scenarios as crosswalk usage or turning conflicts at intersections are evaluated similarly throughout and are better represented by the framework.

However, a difference can be seen not only between different scenarios at the same location but also between infrastructures. In contrast to the classical intersection \textit{Frankenberg}, interaction on straight traffic routes with increased speed is rated higher by participants. In addition, the deviations between raters are smaller at the intersections than at the roundabout (see Fig.~\ref{fig:pearson_neuweiler}). Despite these differences, the interaction score can accurately reflect the participants' opinions. Even at roundabouts, where the opinions of the test subjects are scattered, the framework fits the average opinion of the participants better than the judgment of an individual.
This can also be confirmed by the fact that the evaluation criteria of the test persons fit those of the framework. The participants were asked an open question about the influences of their interaction scoring retrospectively. Most of them mention predicted conflicts (67 percent), the number of surrounding vehicles (67 percent) and compliance with traffic rules (29 percent) as the most important factors. Speed, complexity, type of road user (17 percent each) and stopping of vehicles (4 percent) are less important for their scoring.

\subsection{Comparison with Existing Approaches}

After the correlation between interaction score and human perception has been shown, a comparison of the interaction score is made with a previously proposed framework for the evaluation of trajectory-based datasets. For this purpose, the so-called \textit{INTERACTION} framework is chosen~\cite{INTERACTION_dataset} because it is the most comprehensive one focusing on the interaction in trajectory datasets from bird-eye's view. Besides this metric approach, there is no other method used for dataset evaluation. Compared to our proposed framework, the INTERACTION scheme uses a more simplistic approach since it focuses only on metrics (WP and $\Delta$TTCP) and does not link them to create a comprehensive analysis of road users. For that reason, this scheme can not be compared to the study in chap.~\ref{sec:interaction_score_study} since the study participant were asked to rate the interaction of road users and not bilateral relations. So, the INTERACTION scheme is only compared to our proposed interaction score within a representative sample of usual intersections. The selected recording (\textit{inD 21}) was taken at the intersection \textit{Frankenberg} (see Fig.~\ref{fig:infrastructure_frankenberg}) and contains 550 trajectories derived from a 15 minute long video. Since WP and $\Delta$TTCP are defined punctually, only the punctual interaction score is compared. As a result, it can be observed, that the interaction score of the new framework detects more interactive vehicles and more interaction pairs. This is mostly caused by the detection of following scenarios missed by the two individual metrics. Additionally, the interaction score describes the scenarios much more in detail by observing its environment comprehensively and combining different relation indicators (Tab.~\ref{table:scope_INTERACTION_comparison}).

\begin{table}[h]
	\caption{Scope of interaction evaluation in record \textit{inD 21}}
	\label{table:scope_INTERACTION_comparison}
	\begin{center}
    	\setlength\tabcolsep{5.2pt}
		\begin{tabular}{c|c|c|c}
			\textbf{Method} & \textbf{Vehicles} & \textbf{Interacting pairs} & \textbf{Detections} \\
			\hline
			Interaction score 	& 407 & 676 & 21528 \\
			INTERACTION scheme	& 389 & 478 & 2642 	\\
		\end{tabular}
	\end{center}
\end{table}

\begin{figure}[thpb]
	\centering
	\subfloat[Comparison Interaction score with $\Delta$TTCP]{
		\includegraphics[width=0.45\linewidth]{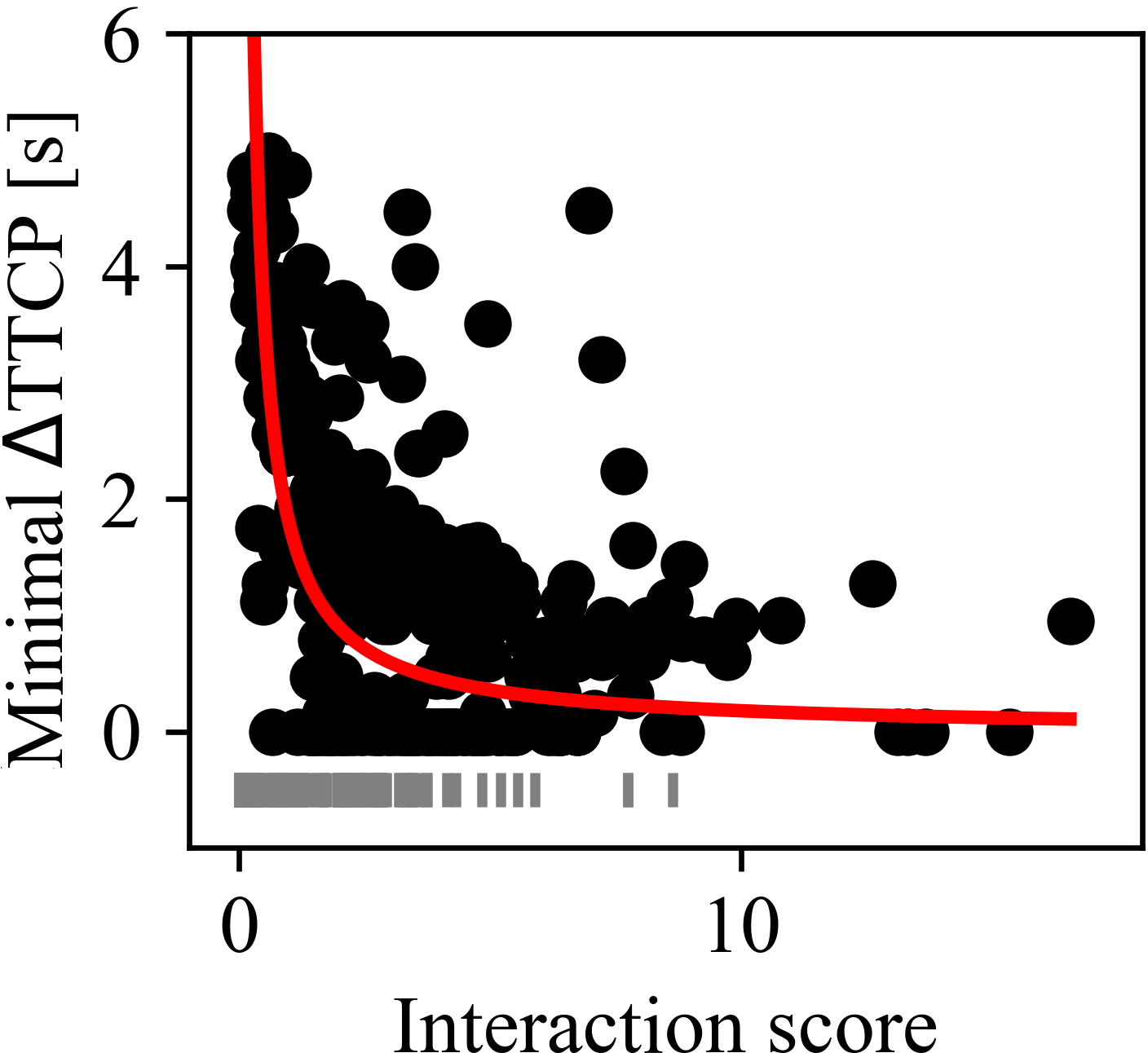}
		\label{fig:comparison_dttcp}
	}
	\hspace{0.03\linewidth}
	\subfloat[Comparison Interaction score with WP]{
		\includegraphics[width=0.45\linewidth]{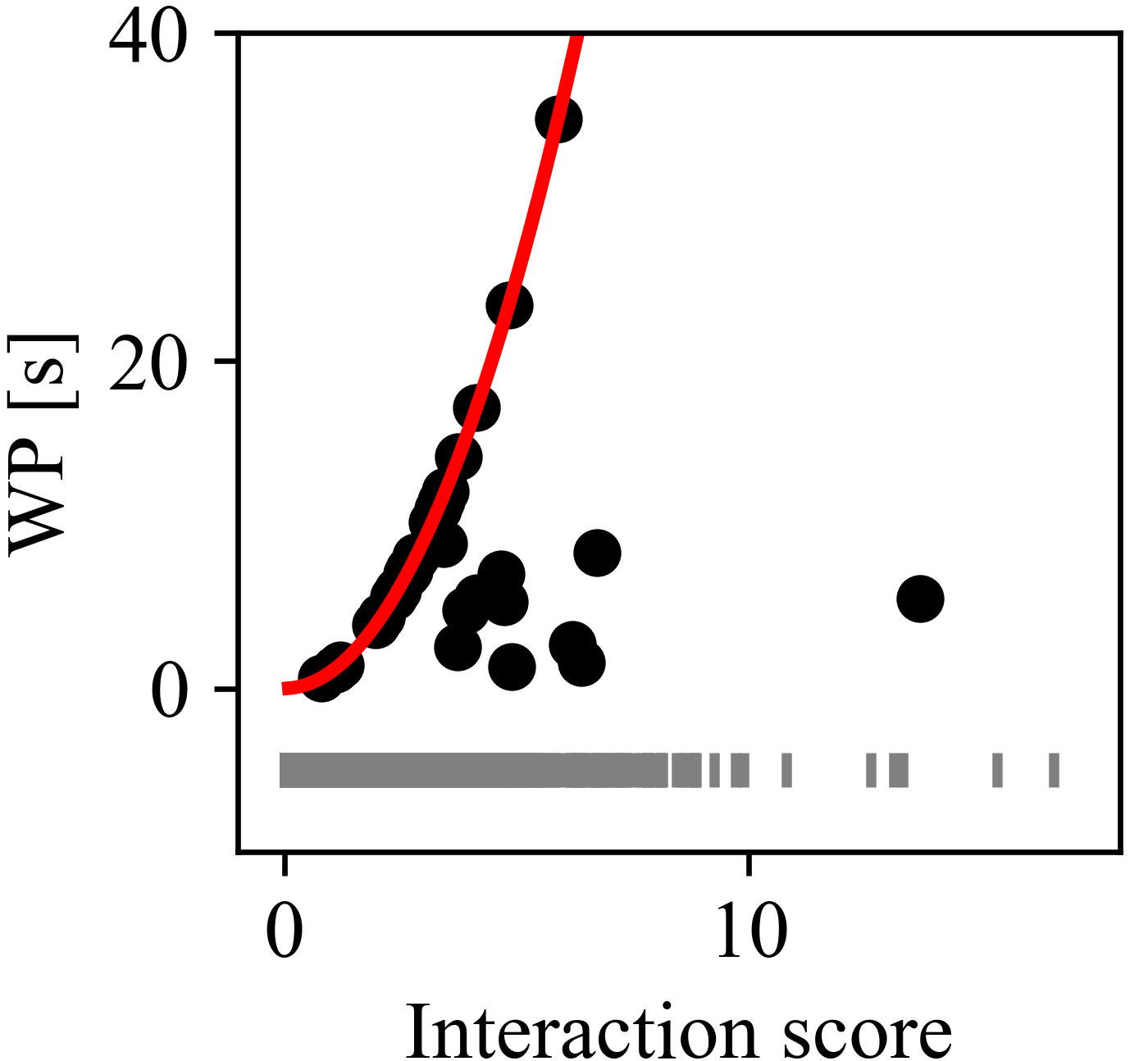}
		\label{fig:comparison_wp}
	}
	\caption{Comparison of the scope of the INTERACTION scheme (y-axis) and new framework (x-axis) including the scoring function of the metric within the interaction framework (red curve, cf. Tab.~\ref{table:detection_scoring}); interactions only found by the interaction score are marked as vertical gray lines}
	\label{fig:comparison_scope}
\end{figure}

Furthermore, the characteristics of the detected interactions differ from those of the INTERACTION scheme. Thus, Fig.~\ref{fig:comparison_scope} shows significant differences between the interaction scores and the metrics translated into the scoring system of the framework. These are due to the more comprehensive evaluation of the environment made by the new framework. On the one hand, large deviations between interaction scores and metrics at low scores are mainly caused by the THW detection and by the number of vehicles involved. So, the interaction score does not oversimplify situations by using just individual metrics, but it takes multiple interacting vehicles into account. On the other hand, the differences at high interaction scores, mostly near misses and collisions, are caused by the TTC and DRAC detections. Because of those additionally used metrics, vehicles are not just evaluated more in-depth, but also the complexity of situations and a broader scope of vehicle relations are considered.

\begin{figure}[thpb]
	\centering
	\includegraphics[width=\linewidth]{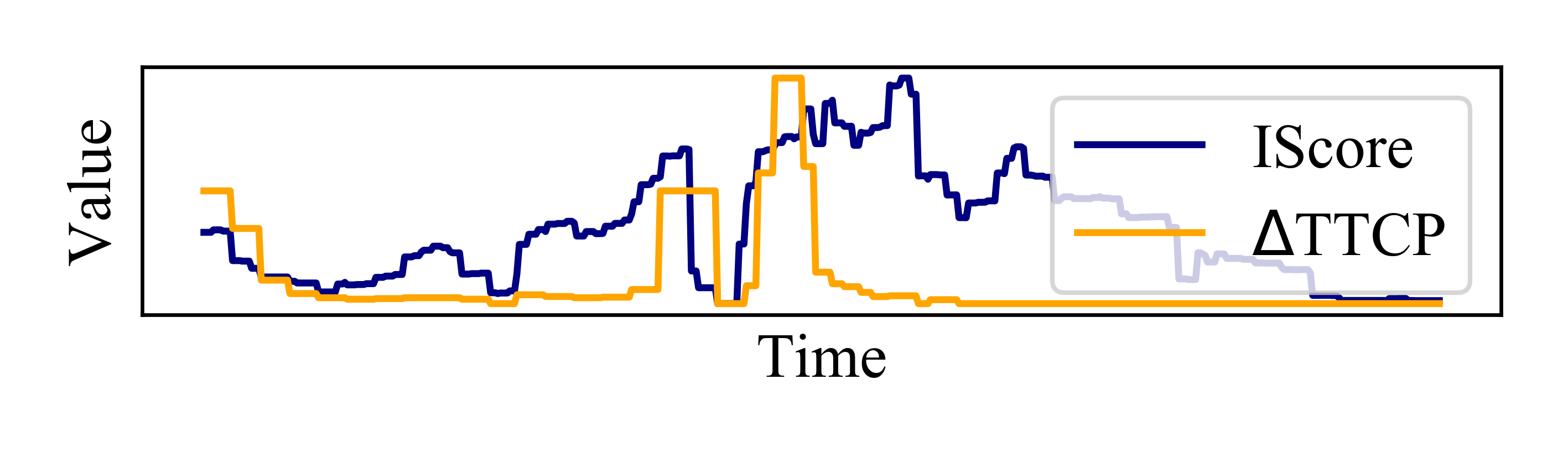}
	\caption{Comparison of $\sum 1 / \Delta$TTCP and punctual interaction score for one vehicle along its trajectory}
	\label{figure:comparison_dttcp_is_vehicle}
\end{figure}

The more detailed evaluation can be illustrated by observing the temporal sequence of the interaction score of an individual vehicle (see Fig.~\ref{figure:comparison_dttcp_is_vehicle}). For this comparison, the $\Delta$TTCP detections are derived in each timestep and their scores are summed up for comparability. That has to be done since the $\Delta$TTCP is only determined for bilateral interaction between road users while the interaction score covers all interactions of a specific road user. The diagram shows not only the more detailed course of the interaction score according to the continuously changing traffic constellations but also a time shift of the global maximum compared to the summed $\Delta$TTCP and additional local maxima. In this example, the shift is a result of the different observation spaces. A large number of interactions cannot be covered by the $\Delta$TTCP, whereas these are included in the interaction score through additional detections. This can cause a different interpretation and leads to a more comprehensive picture. 
While the scores may partially correlate depending on the infrastructure, the shift and score in the last section of the trajectory are caused by a preceding object that is not recognized by the $\Delta$TTCP in this specific example. However, this detected road user in front of the observed vehicle limits its freedom of movement and has to be considered in a comprehensive interaction scoring. So, the maximum of the interaction score is more accurate and fits the human perception.

As shown, the new framework considers the environment especially in complex scenarios more comprehensively and allows a detailed evaluation of the interaction along the road user trajectories. That mostly includes following scenarios as well as the combination of different relations with multiple traffic participants. These often occur when there is an accumulation of road users and various regulatory elements. Relatively similar coverage and evaluation are reached for simpler traffic scenarios such as the interaction of only two traffic participants at an intersection.
A similar conclusion applies to the rating of entire datasets because it can be seen as a sum of its trajectories. While a summation discretized into fixed time gap intervals is used in~\cite{INTERACTION_dataset} for $\Delta$TTCP, the new framework provides a single value to compare a dataset with others. The hierarchical structure of the new method not only makes a comparison simpler, but also provides detailed information along with several layers with different degrees of abstraction that can be used in a much more detailed way than the compared method using two single metrics.

\subsection{Utilization of the Framework}

\begin{figure}[thpb]
	\centering
	\subfloat[Anormal behavior of a car $\mbox{\text{(orange)}}$ passing street, crosswalk and walkway]{
		\includegraphics[width=0.45\linewidth]{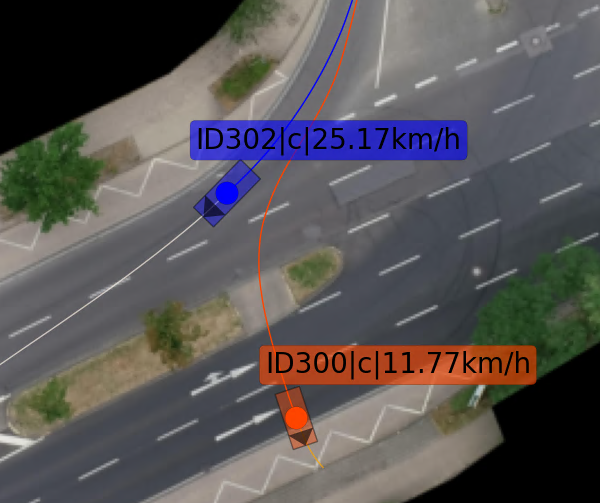}
		\label{fig:buggy_anomaly}
	}
	\hspace{0.03\linewidth}
	\subfloat[Flight route of a bird (orange) detected as a trajectory of a pedestrian]{
		\includegraphics[width=0.45\linewidth]{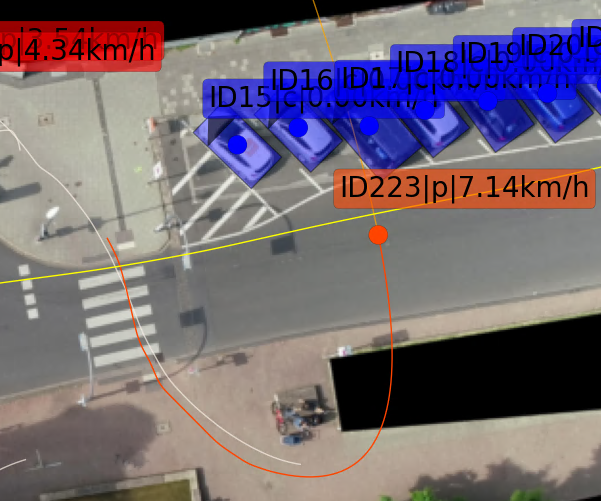}
		\label{fig:bird_as_pedestrian}
	}
	\caption{Examples for highly anomal scored situations.}
	\label{fig:anomaly_situations}
\end{figure}

Beyond the interaction score, the framework includes two other scores to support the user to find anomalies and relevant situations in a dataset. These functions are not considered in previous dataset evaluations. Therefore, they can not be compared to the INTERACTION- or other schemes since the target to score anomaly and relevance in the scope of vehicles, regions and datasets is unique. However, these scores are essential for a comprehensive assessment within the framework. On the lowest level, the punctually defined anomaly score offers two decisive advantages: overview over unusual situations and comprehensive error detection. Since the anomaly score is based on all fourteen detection types rather than just a few like the interaction score, it gives a more comprehensive view of the traffic situation. This is particularly clear at the punctual level. The highest anomaly scores contain unusual behavior patterns (see Fig.~\ref{fig:buggy_anomaly}), which are highly relevant in the context of safeguarding automated driving functions. The scores are normally significantly impacted by the clustering and infrastructure-related detections because they often occur together. For example, an unusual trajectory often correlates with locally unusual behavior. Nevertheless, the situations found are not always unusual subjectively. If just a single vehicle is driving into a parking bay, it is highly abnormal within the context of the record. If a human anticipates the driver's behavior and includes this knowledge into its evaluation of the situation, the situation might seem usual for him. Anyways, the scoring makes sense from an unbiased point of view when considering the actions within the given record. This gap between objective and subjective evaluation is not necessarily bad since it shows the characteristic of the dataset and creates awareness for rare scenarios. Of course, it is up to the user to subsequently filter out unwanted scenarios such as parking vehicles, considering the given use case.
In addition to behavior detection, errors can be identified as anomalies. Whether it is a bird that has been assigned as a pedestrian (see Fig.~\ref{fig:bird_as_pedestrian}) or a wrongly detected vehicle.

Applying the framework on the \textit{inD} dataset and filtering the data for the most unusual scenarios one out of ten scenarios is caused by false image object detection. Two others include vehicles driving on designated pedestrian crosswalks perpendicular to the normal driving direction (see Fig.~\ref{fig:buggy_anomaly}). One vehicle does a U-turn, two performing parking maneuvers, another vehicle drives in the middle of two lanes and in a next situation one does a left turn resulting in driving in the middle of two lanes, too. Furthermore, two other situations show pedestrians crossing large roads without using designated crosswalks. So, as already shown above, the anomaly score mainly highlights exceptional situations, but also includes scenarios that are not rare generally but in the observed data.
Having a look at the ten most interactive scenarios gives a less diverse picture. Thereby, four situations are cluttered and contain each more than seven interacting road users. Two situations are assigned to wrong detected collision and two more to near misses. The last two are pedestrian-related and include disembarking from a bus and bypassing a construction site. Whereas the groups of pedestrians might be less relevant for automated driving functions, especially complex and near-miss scenarios can help to validate and optimize those functions heavily.
After having a look at anomaly and interaction, the relevance score is elaborated as the product of it.
Although it is composed of both scores, the highest-rated interaction or anomaly situations do not necessarily lead to high relevance scores. Instead, a high relevance score is mainly driven by the interplay of multiple vehicles and complex situations.
Due to the multiplication of predetermined scores, its values are subject to a higher error tolerance than the uncombined ones: Because of the higher sensitivity to accidents, errors in vehicle detection can have a big impact on the relevance score. If vehicle boxes overlap with others or are wrongly assigned to regions, the detection may lead to a virtual collision or the occurrence of other detections.
Therefore, individual situations can require manual review due to errors in datasets. On more abstract levels, the proposed approaches are more resistant to dataset errors or weaknesses of single metrics so that they can provide reliable values for records.
As explained below especially the relevance score can help to find errors in datasets. So, four out of ten of the most relevant situations of the inD dataset are caused by wrong detections and pseudo collisions, three are related to interesting parking maneuvers and the others can be assigned to complex scenarios on intersections, near misses and a vehicle on a sidewalk. Depending on the use case, those scenarios are more or less relevant. Although the wrong detections might seem useless, they are especially useful for publishers of datasets to check for the quality of their data. Besides that, the punctual relevance score, as well as anomaly and interaction, show different and diverse situations helping a user to apply a variety of situations to driving functions and getting an overview about interesting situations in the dataset.

\begin{figure}[thpb]
	\centering
	\includegraphics[width=0.9\linewidth]{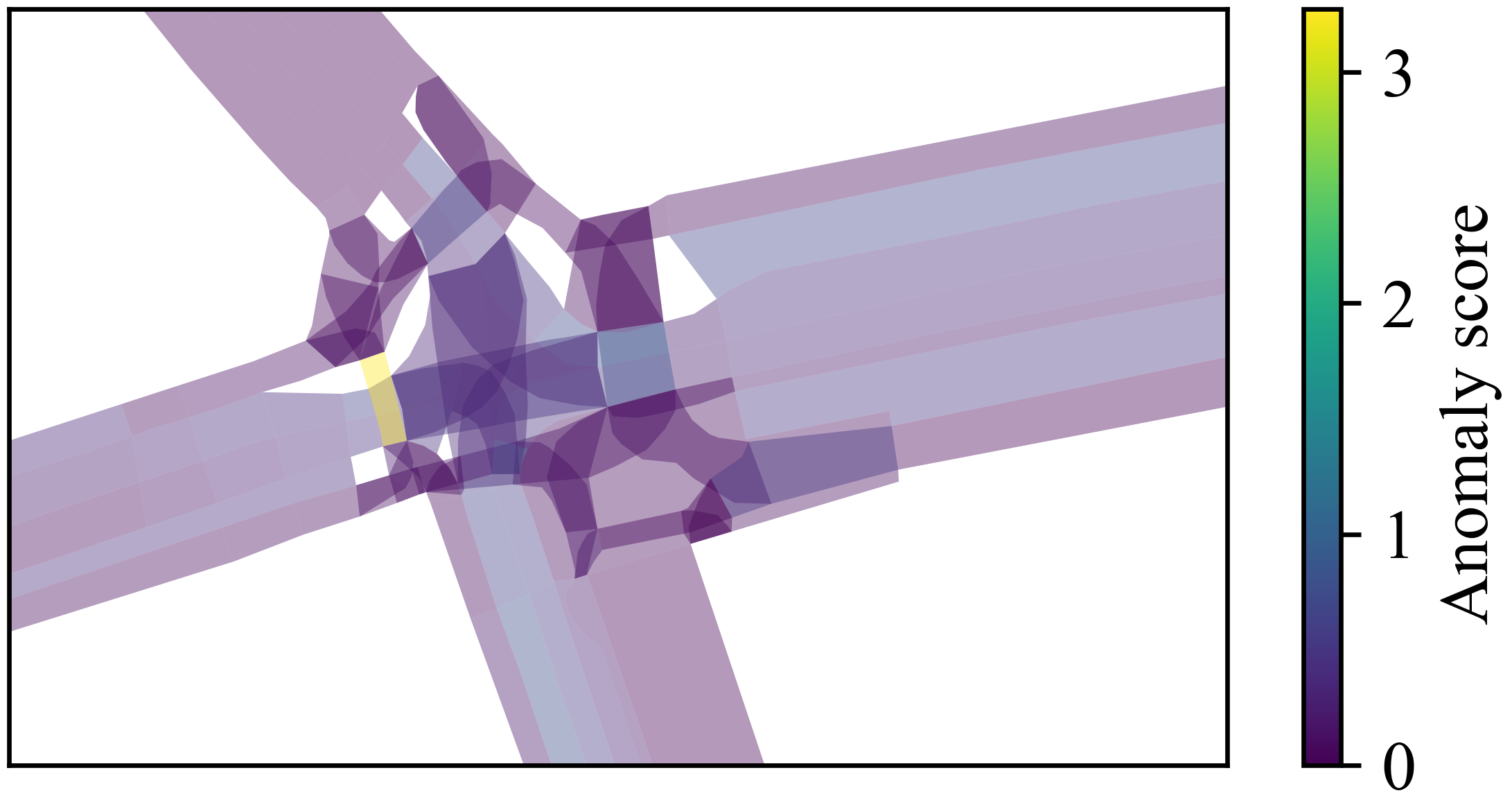}
	\caption{Heatmap of anomaly region scores of \textit{inD} 21}
	\label{fig:anomaly_map}
\end{figure}

Beyond the detection of individual situations, more abstract statements can be made with all three scores. Information about the general traffic situation can be obtained by looking at the regional distribution of scores. This distribution not only makes abnormal behavior visible in each location, but also allows conclusions to be drawn about infrastructure elements. In Fig.~\ref{fig:anomaly_map}, it is evident that the non-signed crosswalk over the left arm of the intersection has a significantly higher anomaly score compared to the signed crosswalk over the right arm. This finding could lead to the hypothesis that regulatory elements prevent anomalous crossings. With this kind of analysis, it is possible to draw conclusions about individual elements and then select infrastructures for future recordings according to the underlying requirements of e.g. scenario-based testing. This selection can be supported not only by the three scores themselves but additionally also by their composition of belonging detection types.
	
\begin{figure*}[t]
	\centering
	\subfloat{
		\includegraphics[width=0.3\linewidth]{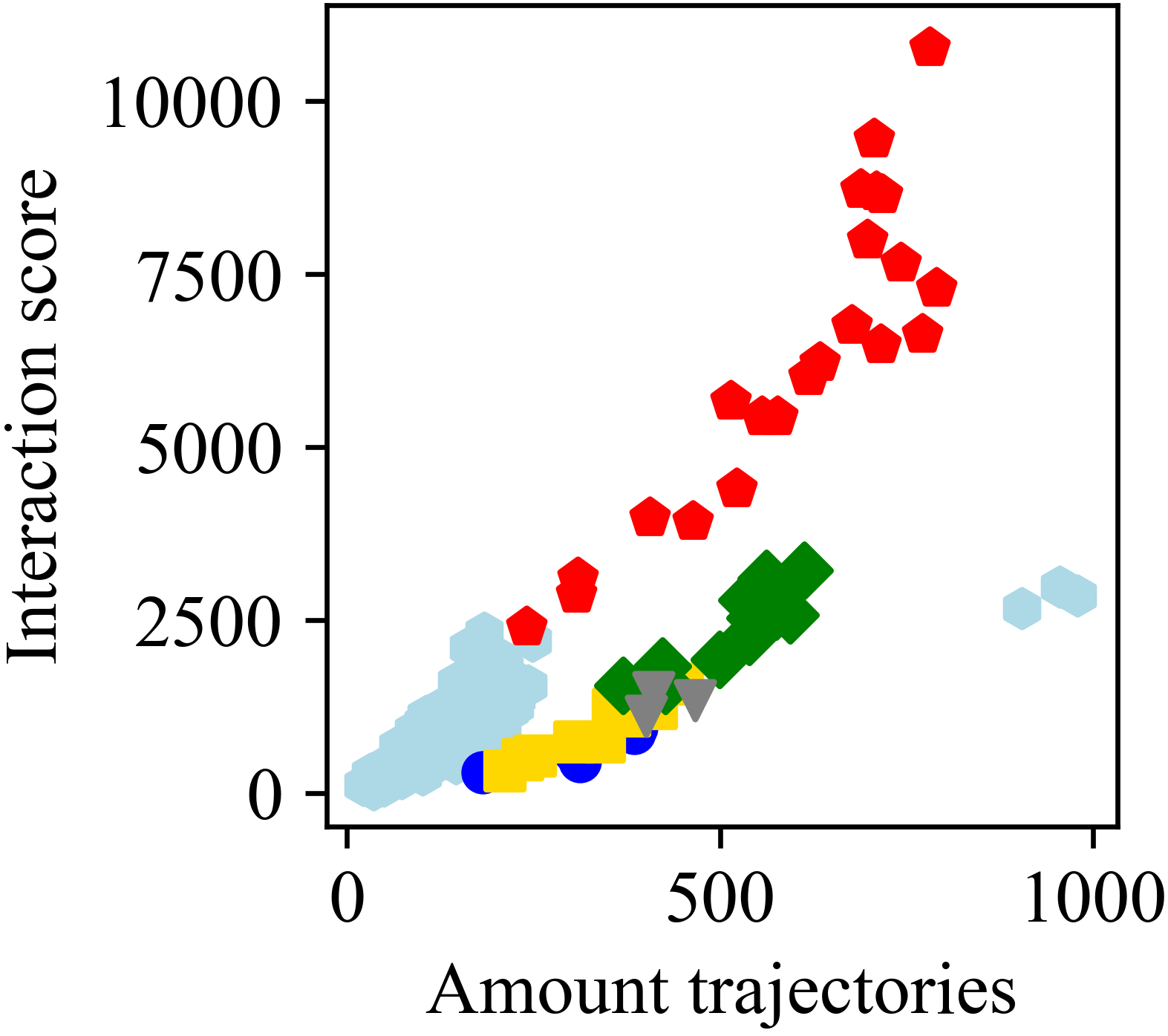}
		\label{fig:dataset_overview_interaction}
	}
	\hspace{0.02\linewidth}
	\subfloat{
		\includegraphics[width=0.3\linewidth]{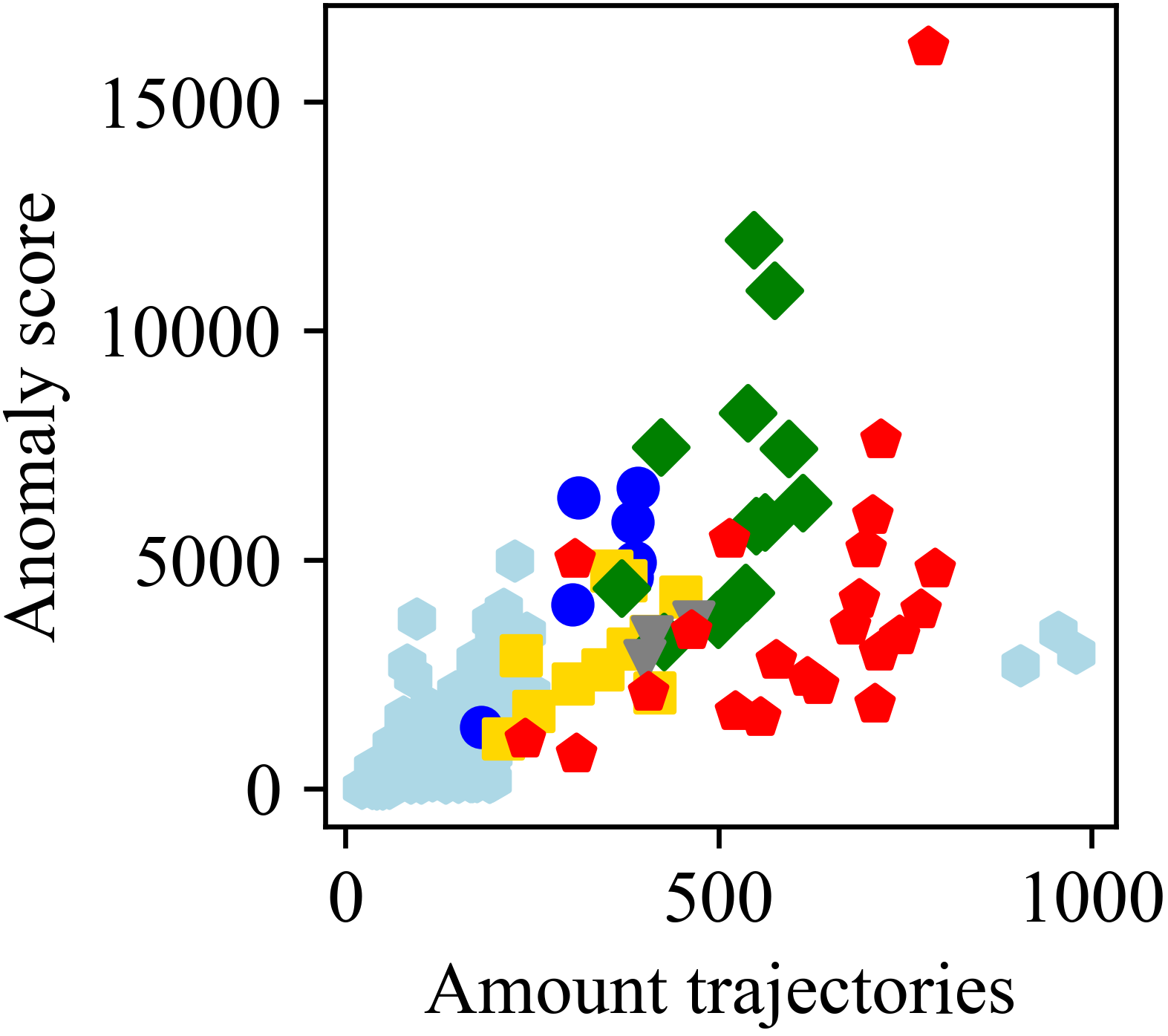}
		\label{fig:dataset_overview_anomaly}
	}
	\hspace{0.02\linewidth}
    	\subfloat{
		\includegraphics[width=0.3\linewidth]{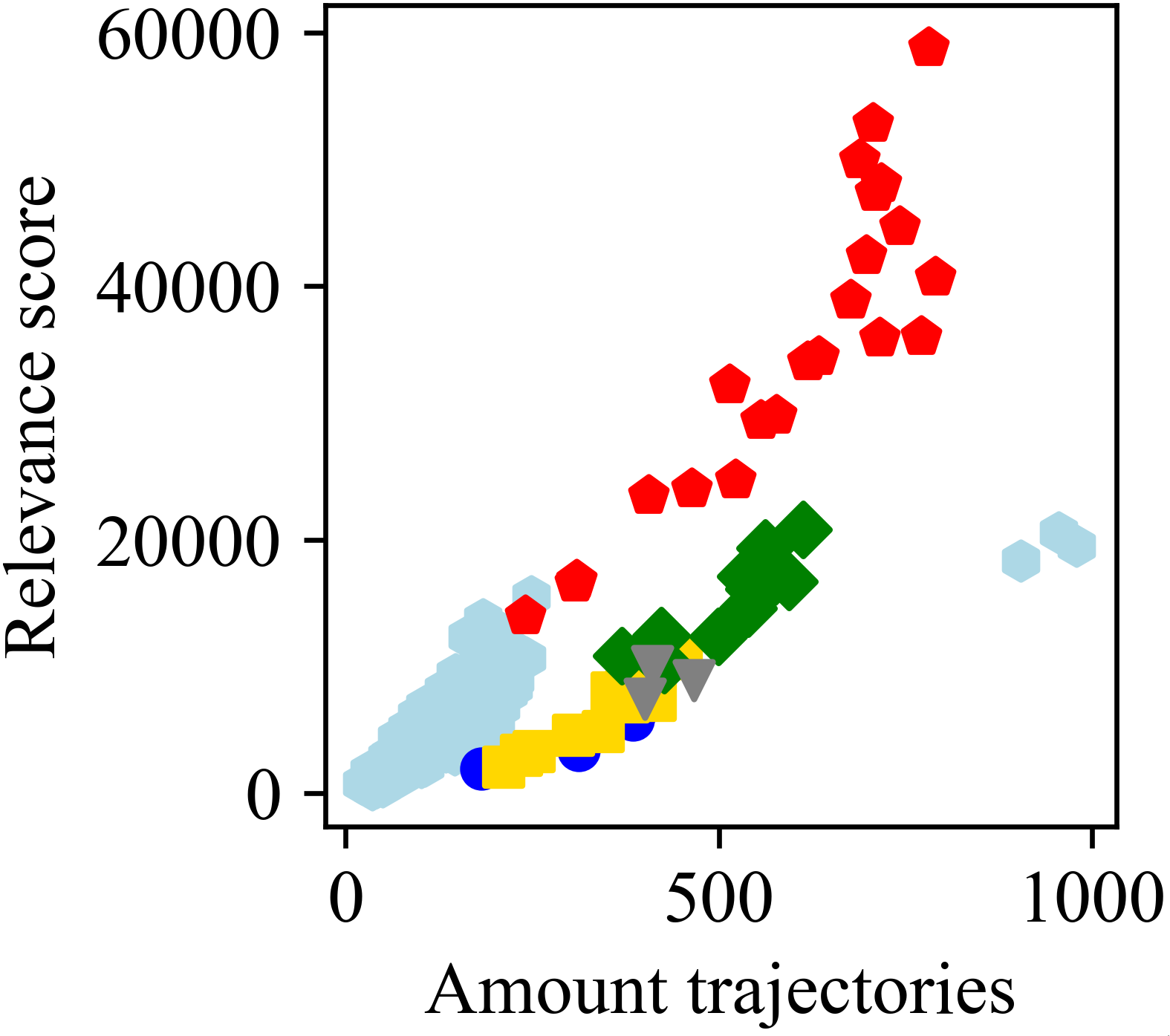}
		\label{fig:dataset_overview_relevance}
	}
	\caption{Comparison of overview scores for urban records of \textit{inD}~\cite{inD}, \textit{rounD} (red pentagons)~\cite{rounD} and \textit{INTERACTION} (light blue hexagons)~\cite{INTERACTION_dataset} with detailed view on \textit{inD} infrastructures (\textit{Aseag}: deep blue circles, \textit{Bendplatz}: yellow squares, \textit{Frankenberg}: green diamonds and \textit{Heckstrasse}: gray triangle)}
	\label{fig:dataset_comparison}
\end{figure*}

Thus, the combination of the three scores also provides an overview of the level of entire trajectory datasets and allows a comparison of them (see Fig.~\ref{fig:dataset_comparison}). Comparing interaction and relevance scores in over 200 recordings, it can be observed, that interaction and relevance are highly linked for individual road users in the given recordings. Because of the high traffic density, unusual behavior causes further interaction with other road users.
While the three scores are similar for different recordings at the same location due to similar road user behavior, they significantly differ between different location sites and different datasets. Across the observed datasets, it can be observed that rounD might suit well for testing complex multi-road user scenarios due to their high interaction. If a user would be more interested in unusual road user behavior, the inD dataset and especially the intersection \textit{Frankenberg} might fit better. This kind of analysis can be used to select appropriate recordings for the individual application and to draw conclusions about the driving behavior in different locations in order to plan future datasets recordings according to one's own needs.

Thus, the framework offers several possibilities for the analysis of trajectory-based trajectory datasets. As the framework utilizes a set of weights for combining the individual scores, use case specific adaptation is possible and necessary for best results. Furthermore, a graphical interface can help both, to get an quick overview as well as analyzing scores quickly and in-depth.

\section{CONCLUSIONS}

In this paper, an automated analysis framework for urban trajectory-based datasets from a bird's eye view was presented. Both, existing and newly derived metrics were combined with approaches of anomaly detection and clustering to take vehicle constellations and driving behavior into account. Based on these fourteen traffic detection types, interaction, anomaly and relevance of the data were determined on different hierarchical levels. Thus, the framework showed a high correlation between the interaction score and human perception of interaction. Mostly, the framework reflects this perception better than asking a single expert. It remains part of future research to find study designs that allow validation also for the anomaly- and relevance score.
The fact of outperforming individual experts and the complete automation of the framework allows different datasets to be compared in-depth. This can be done in more detail and with more clarity than with existing approaches. 
The proposed method brings not only a benefit for users of datasets but also for publishers, as they can draw conclusions for future recordings.
Although the evaluation framework has been optimized for urban scenarios in its current form, the modular design of the detections allows adaptations to further use cases so that the framework can be extended. 
 
%%%%%%%%%%%%%%%%%%%%%%%%%%%%%%%%%%%%%%%%%%%%%%%%%%%%%%%%%%%%%%%%%%%%%%%%%%%%%%%%

\end{document}